\documentclass[sigconf]{acmart}
\AtBeginDocument{%
  }

\copyrightyear{2026}
\acmYear{2026}
\setcopyright{cc}
\setcctype{by}
\acmConference[MM '26]{Proceedings of the 34th ACM International Conference on Multimedia}{November 10--14, 2026}{Rio de Janeiro, Brazil}
\acmBooktitle{Proceedings of the 34th ACM International Conference on Multimedia (MM '26), November 10--14, 2026, Rio de Janeiro, Brazil}
\acmDOI{10.1145/3767308.3836564}
\acmISBN{979-8-4007-2213-4/2026/11}

\usepackage{subcaption}
\usepackage{siunitx}
\usepackage{multirow}
\usepackage{makecell}
\usepackage{arydshln}
\usepackage{tabularx}
\usepackage{bbding}
\usepackage{pifont}
\usepackage{mathtools}
\usepackage{algorithm}
\usepackage{algorithmic}

\usepackage[capitalize,noabbrev]{cleveref}

\definecolor{mygreen}{RGB}{0, 150, 0}
\definecolor{myred}{RGB}{200, 0, 0}

\theoremstyle{definition}

\theoremstyle{remark}

\definecolor{lowred}{RGB}{238,18,137}
\definecolor{dpluscolor}{RGB}{77,140,80}

\newcommand{\dplus}[1]{\fontsize{6pt}{0.1em}\selectfont (\textbf{\textcolor{dpluscolor}{#1}})}

\begin{document}

\title{Multi-Branch Policy Optimization for Multimodal Large Language Models}


\settopmatter{authorsperrow=4}

\title{Multi-Branch Policy Optimization for Multimodal Large Language Models}


\settopmatter{authorsperrow=4}

\title{Multi-Branch Policy Optimization for Multimodal Large Language Models}


\author{Shuai Lyu}
\authornote{These authors contributed equally.}
\affiliation{%
  \institution{Beijing University of Posts and Telecommunications}
  \city{Beijing}
  \country{China}}
\email{Lxb_savior@bupt.edu.cn}

\author{Yuning Gong}
\authornotemark[1]
\affiliation{%
  \institution{Sichuan University}
  \city{Chengdu}
  \country{China}}
\email{2021323040003@stu.scu.edu.cn}

\author{Ruiling Gao}
\affiliation{%
  \institution{Shanghai University}
  \city{Shanghai}
  \country{China}}
\email{ruilinggao@shu.edu.cn}

\author{Xiaoran Shang}
\affiliation{%
  \institution{Beijing University of Posts and Telecommunications}
  \city{Beijing}
  \country{China}}
\email{sxr15@bupt.edu.cn}

\author{Zhonghong Ou}
\affiliation{%
  \institution{Beijing University of Posts and Telecommunications}
  \city{Beijing}
  \country{China}}
\email{zhonghong.ou@bupt.edu.cn}

\author{Ping Zong}
\affiliation{%
  \institution{Beijing University of Posts and Telecommunications}
  \city{Beijing}
  \country{China}}
\email{2010919530@bupt.cn}

\author{Yifan Zhu}
\affiliation{%
  \institution{Beijing University of Posts and Telecommunications}
  \city{Beijing}
  \country{China}}
\email{yifan_zhu@bupt.edu.cn}

\author{Yuan Sun}
\affiliation{%
  \institution{Sichuan University}
  \city{Chengdu}
  \country{China}}
\email{sunyuan_work@163.com}

\author{Yang Qin}
\authornote{Corresponding authors.}
\affiliation{%
  \institution{Huawei Technologies Ltd.}
  \city{Shenzhen}
  \country{China}}
\email{qinyang.gm@gmail.com}

\author{Peng Hu}
\authornotemark[2]
\affiliation{%
  \institution{Sichuan University}
  \city{Chengdu}
  \country{China}}
\email{penghu.ml@gmail.com}

\renewcommand{\shortauthors}{Lyu, Gong, et al.}

\begin{abstract}
Group-based reinforcement learning methods for multimodal large language models typically rely on trajectory-level credit assignment that applies a single advantage to all tokens in a response. However, multimodal reasoning involves substantially higher perceptual uncertainty than text-only settings, the model must repeatedly re-examine visual information to verify intermediate interpretations, and different visual groundings can lead to divergent reasoning paths, making such uniform credit assignment particularly inadequate and causing relative advantages to progressively degenerate toward zero. To address these challenges, we propose Multi-Branch Policy Optimization (MBPO), a tree-based framework that constructs reasoning trees at vision-language decision boundaries, enabling sibling branches to explore diverse visual hypotheses and assigning segment-level credit through branch-relative advantages. We further introduce a temporal replay buffer to reuse informative segments while controlling policy staleness. Experiments on several multimodal reasoning benchmarks show that MBPO outperforms representative baselines, improving both learning signal quality and optimization efficiency. The code is publicly available at \url{https://github.com/ShuaiLyu0110/MBPO}.
\end{abstract}




\maketitle


\section{Introduction}

\begin{figure}[bht!]
  \centering
  \begin{subfigure}{1.0\columnwidth}
    \centering
    \includegraphics[width=1\linewidth]{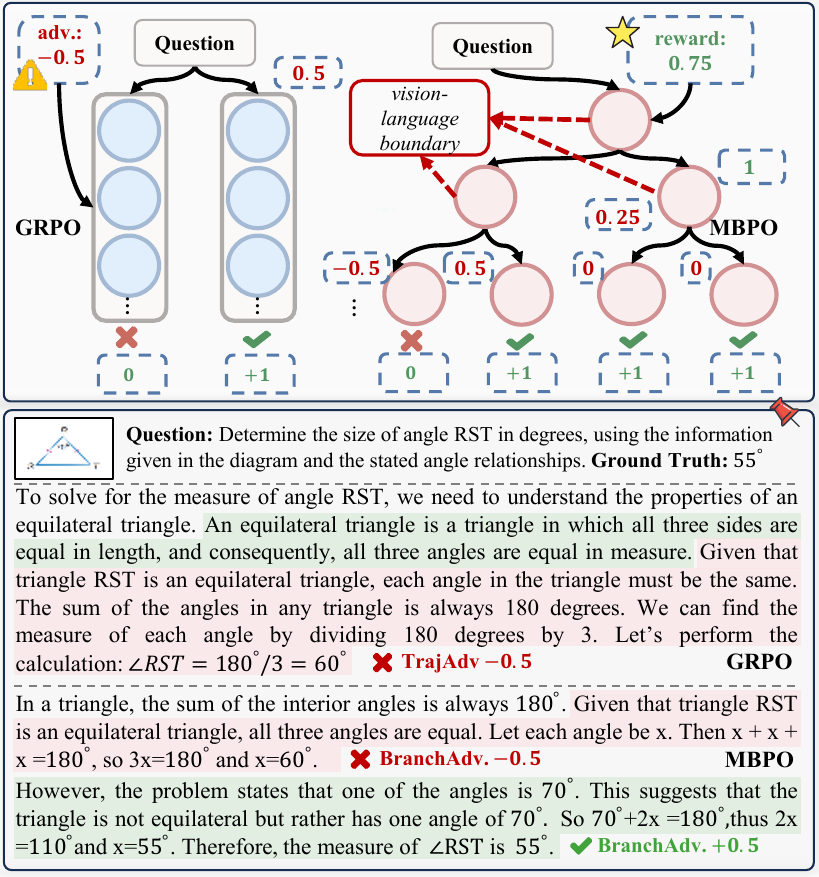}
  \end{subfigure}
  \caption{
  The difference between MBPO and GRPO in terms of reward and advantage propagation. Unlike the \emph{one-size-fits-all} strategy that uses a single trajectory-level advantage for the full response, our branch-level MBPO assigns a local negative advantage to the incorrect branch and a positive one to the correct branch, enabling more accurate global advantage assignment.  This allows for fine-grained reward while mitigating relative advantage collapse.
  }
  \label{fig:f1}
  \vspace{-5mm}
\end{figure}

With the recent success of reinforcement learning (RL)~\cite{deng2025atom}, training paradigms have gradually shifted from purely supervised fine-tuning to outcome-driven optimization for Multimodal Large Language Models (MLLMs). Group Relative Policy Optimization~\cite{deepseek-math} (GRPO), as a representative RL method, assigns outcome-based advantages to generated reasoning trajectories, providing a simple, verifiable training signal for reasoning in MLLMs. However, existing GRPO-style methods treat multimodal reasoning as a single linear token sequence and apply uniform trajectory-level credit assignment, ignoring that only a few perceptual or reasoning segments are causally decisive. Compared with text-only reasoning, multimodal tasks introduce additional decision points: visual grounding, cross-modal alignment, and perceptual integration, that increase both the number of branching opportunities and the overall chain length, amplifying credit ambiguity and making trajectory-level credit assignment particularly inadequate.

\textbf{A distinctive challenge of multimodal reasoning is the inherent visual uncertainty of images}: a single image region can admit multiple plausible interpretations, and the downstream reasoning chain is highly sensitive to whichever interpretation the model commits to at each step. The model must therefore repeatedly re-examine visual information to verify its perceptual understanding, and different re-examinations can lead to divergent yet individually coherent reasoning paths. This perceptual diversity calls for a structured exploration mechanism that can systematically branch at vision-language decision points and compare sibling hypotheses grounded in different visual interpretations.

\begin{figure}[t]
  \centering
  \begin{subfigure}{1.0\columnwidth}
    \centering
    \includegraphics[width=1\linewidth]{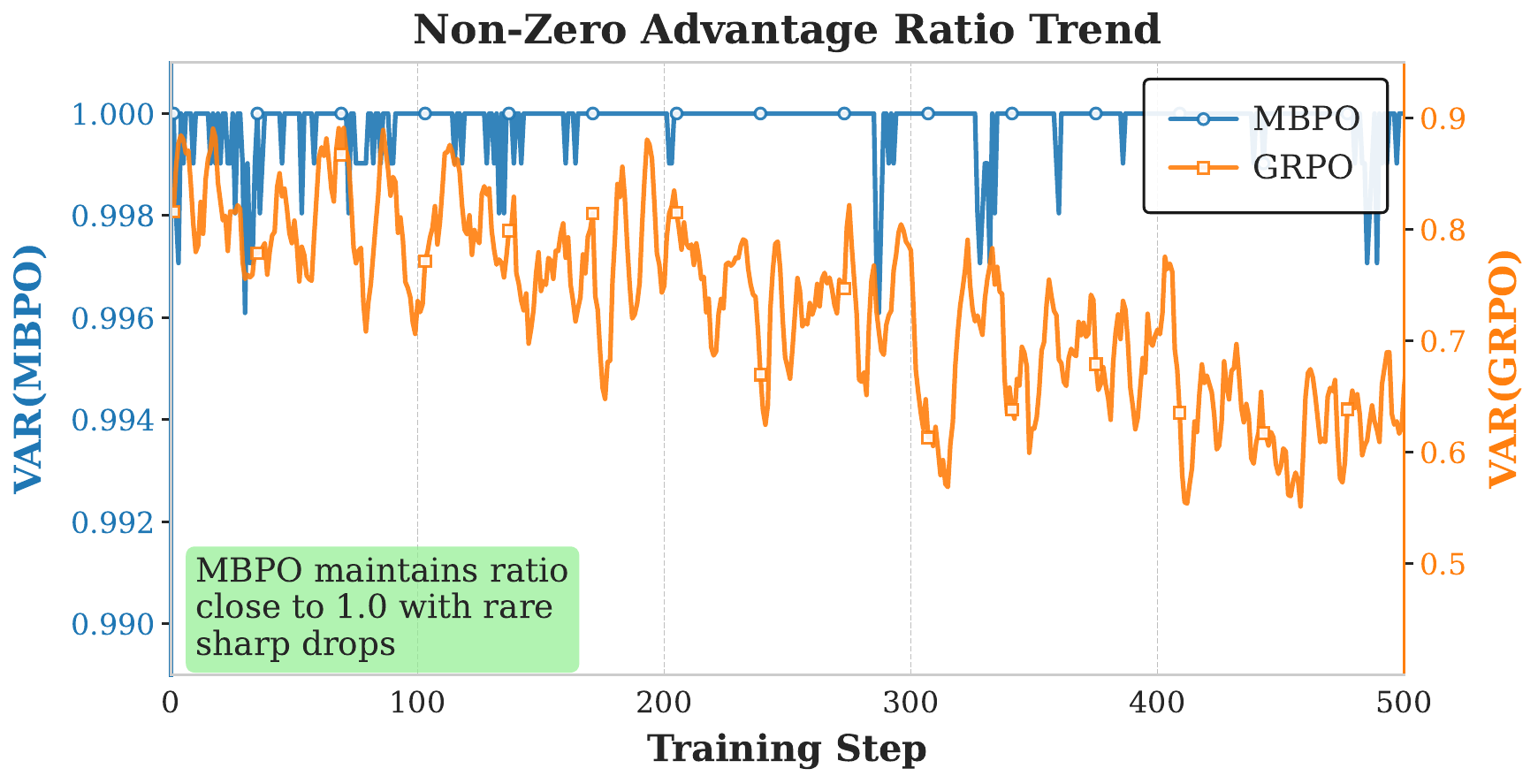}
  \end{subfigure}
    \caption{Relative advantage degeneration during training on Geo3K with Qwen2.5-VL-3B-Instruct. MBPO remains stable, while GRPO shows a decreasing trend over training steps.}
  \label{fig:f2}
\end{figure}

This property also exposes a limitation of trajectory-level policy optimization. As illustrated in \Cref{fig:f1}, a response may contain an incorrect intermediate segment followed by a correction, yet GRPO assigns the same trajectory-level advantage to all tokens, blurring the learning signal at critical decision points. Over time, this credit confusion triggers relative advantage degeneration, where group-relative advantages collapse toward zero as rollouts become more similar.
We track this degeneration using the valid advantage ratio (VAR; \cref{eq:VAR}) in \Cref{fig:f2}, following R1-ShareVL \cite{yao2025r1} and Shuffle-R1 \cite{zhu2025shuffle}, and observe a significant and consistent decline.

To address these challenges, we propose Multi-Branch Policy Optimization (MBPO), which improves credit assignment through branch-level advantage estimation. Tree search is a natural fit for exploring the perceptual uncertainty of multimodal reasoning: sibling branches pursue distinct visual hypotheses from a common prefix, allowing the optimizer to directly compare alternative visual groundings rather than averaging over a single trajectory (\Cref{fig:f1}). We further introduce a temporal replay buffer that retains informative segments to reduce policy drift. As shown in \Cref{fig:f2}, MBPO maintains a more stable non-zero advantage proportion than GRPO throughout training. Evaluated on Geometry3K and MMK12, MBPO generalizes to six out-of-domain benchmarks and consistently outperforms RL baselines including GRPO and DAPO \cite{yu2025dapo}, as well as Shuffle-R1 and MM-Eureka-Qwen-7B when trained on MMRL18K. 

Our main contributions are summarized as follows:
\begin{itemize}
    \item We propose \textbf{Multi-Branch Policy Optimization (MBPO)}, a tree-structured RL framework that leverages the inherent visual diversity in multimodal reasoning to construct reasoning trees at vision-language decision boundaries, enabling branch-level credit assignment via \emph{sibling-relative} advantages.

    \item We introduce a \textbf{temporal replay buffer} with question-balanced sampling to reuse informative segments while limiting policy staleness.

    \item We demonstrate consistent gains over strong RL baselines on Geometry3K and MMK12, and improved out-of-domain generalization across six multimodal reasoning benchmarks.
\end{itemize}

\section{Related Work}

\begin{figure*}[bht!]
  \centering
  \includegraphics[width=\textwidth]{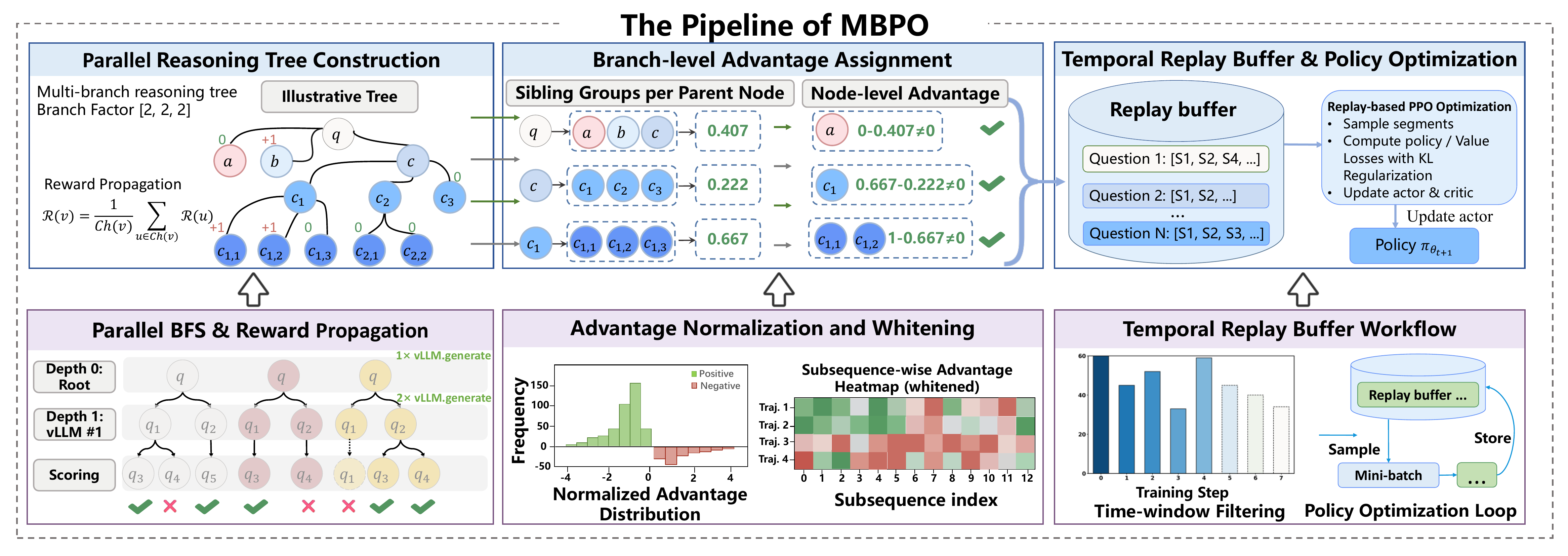}
  \caption{
The illustration of MBPO. The figure illustrates how MBPO uses tree-structured rollouts to share reasoning branches and compute branch-level advantages. It shows the training with advantage normalization and a temporal replay buffer.}
  \label{fig:f3}
\end{figure*}

\subsection{RL for MLLMs Reasoning}
With the development of multimodal communities~\cite{qin2026robust,qin2025human,fengmultimodal}, RL has become an important tool for improving the reasoning ability of MLLMs. Recent studies extend RL from text-only reasoning to MLLMs and
downstream vision tasks. For example, R1-VL \cite{zhang2025r1} adapts GRPO to MLLMs and introduces StepGRPO with stepwise reasoning rewards, while R1-ShareVL \cite{yao2025r1} proposes Share-GRPO with question expansion, trajectory sharing, and hierarchical advantage estimation to alleviate sparse rewards and advantage vanishing. Shuffle-R1 \cite{zhu2025shuffle} studies the training dynamics of GRPO and uses pairwise trajectory sampling and advantage-based batch shuffle to reduce advantage collapsing and rollout silencing. DAPO \cite{yu2025dapo} introduces a token-level policy gradient loss that modifies the loss attribution at the token dimension of a trajectory to stabilize optimization. Unlike these methods, we represent each candidate solution as a multimodal reasoning tree and assign relative advantages among sibling branches, providing a principled way to exploit  tree structure in RL for multimodal reasoning.

\subsection{Tree-based Policy Optimization}
Several concurrent works also integrate tree structures into RL training for LLMs. TreePO \cite{li2025treepo} focuses on heuristic tree rollout with dynamic divergence and KV-cache reuse to improve sampling efficiency. TreeRPO \cite{yang2025treerpo} estimates step-level reward expectations through tree sampling to construct dense process rewards. TreeRL \cite{hou2025treerl} introduces uncertainty-driven on-policy tree search that branches from high-uncertainty steps for intermediate supervision. Notably, existing tree-based methods operate in the text-only domain and define branches solely through textual cues. In contrast, MBPO extends tree-structured policy optimization to multimodal reasoning through vision-language-driven branching and assigns sibling-relative advantages under shared prefixes for segment-level credit assignment. Unlike TreePO, TreeRPO, and DAPO, which respectively target sampling efficiency, step-level reward estimation, and token-level loss aggregation, MBPO addresses both credit ambiguity and relative advantage degeneration. 

\section{Methodology}
\label{sec:method}
This section introduces three core components of MBPO: (1) reasoning tree construction via breadth-first search (BFS) for efficient exploration, (2) branch-level advantage assignment and reward propagation for precise credit assignment, and (3) a temporal replay buffer to limit the reuse of outdated reasoning segments.

\subsection{Preliminaries}
\label{sec:prelim}
\paragraph{Relative Advantage Degeneration.} We characterize a phenomenon referred to as relative advantage degeneration: as training progresses, relative advantages among model responses gradually flatten, leaving fewer responses with meaningful learning signals. To measure its severity, we define the valid advantage ratio as:
\begin{equation}
\mathrm{VAR} = \frac{1}{K} \sum_{i=1}^{K} \mathbb{I}(|A_i| > 0),
\label{eq:VAR}
\end{equation}
where $\mathbb{I}(\cdot)$ is the indicator function. $\mathrm{VAR}$ measures the proportion of non-zero advantages. Under relative advantage degeneration, $\mathrm{VAR}$ declines, resulting in fewer informative updates. To address this, MBPO prioritizes updates on branches where $|A|>0$, maintaining a high valid advantage ratio during reasoning.

\paragraph{Problem Formulation.} A multimodal reasoning task is defined as follows: Given input $x = (I, Q) \in \mathcal{V} \times \mathcal{T}$, where $I$ is an image or a set of images from the visual space $\mathcal{V}$ and $Q$ is a text query from the text space $\mathcal{T}$. The response space is denoted as $\mathcal{Y}$, and the response $y$ should correctly answer the question and provide a clear reasoning process.

\subsection{Parallel Reasoning Tree Construction}
We model reasoning as a tree of parallel rollouts to enable branch-level advantage estimation among sibling branches. To align branching with reasoning boundaries, MBPO uses a fixed token budget with adaptive detection of vision-language decision points marked by the special token \texttt{<look>}. When \texttt{<look>} is detected, the preceding segment becomes a branching node where sibling branches explore different visually grounded continuations. Otherwise, MBPO uses fixed-length segmentation. This balances semantic branching with stable sibling comparison (Section~\ref{sec:branch_seg}).
\begin{algorithm}[h!]
\small
\caption{Parallel BFS Tree Construction for MBPO}
\label{alg:bfs-tree}
\begin{algorithmic}[1]
\STATE \textbf{Input:} prompts $\{x_i\}_{i=1}^{B}$, policy $\pi_\theta$, branch factors $[K_1,\dots,K_D]$, step length $M$, context limit $L$.
\STATE \textbf{Output:} tree $\mathcal{T}$ and training segments $\mathcal{S}$.

\STATE $\mathcal{L}\leftarrow\emptyset$, $\textsc{Current}\leftarrow\{x_1,\dots,x_B\}$

\FOR{$d=1$ \textbf{to} $D$}
    \STATE Split $\textsc{Current}$ into $\textsc{Expand}$ and $\textsc{Done}$; update $\mathcal{L}\leftarrow\mathcal{L}\cup\textsc{Done}$

    \STATE \textcolor{blue}{\textit{// Set generation budget at depth $d$}}
    \FOR{each $n\in\textsc{Expand}$}
        \STATE $m_n\leftarrow L-|\mathrm{context}(n)|$ if $d=D$, else $\min(M,\;L-|\mathrm{context}(n)|)$
    \ENDFOR

    \STATE \textcolor{blue}{\textit{// Expand all current nodes in one parallel call}}
    \STATE $\textsc{Outputs}\leftarrow\textsc{BatchGenerate}(\textsc{Expand};\,\pi_\theta,\,n=K_d,\,\{m_n\})$, $\textsc{Next}\leftarrow\emptyset$

    \FOR{each $(n,O)\in\textsc{zip}(\textsc{Expand},\textsc{Outputs})$}
        \FOR{each sample $s\in O$}
            \STATE $\mathbf{r}\leftarrow s.\text{tokens}$, $\textit{fin}\leftarrow (s.\text{finish}\neq\texttt{length})$
            \STATE \textcolor{blue}{\textit{// Truncate at vision-language boundary if detected}}
            \IF{$d<D$ and a vision-language boundary marker appears at position $p$ in $\mathbf{r}$}
                \STATE $\mathbf{r}\leftarrow\mathbf{r}_{1:p}$, $\textit{fin}\leftarrow\textsc{false}$
            \ENDIF
            \STATE Create child $c$ and add edge $(n,c)$
            \STATE Add $c$ to $\mathcal{L}$ if $d=D$ or $\textit{fin}$, else to $\textsc{Next}$
        \ENDFOR
    \ENDFOR

    \STATE $\textsc{Current}\leftarrow\textsc{Next}$
\ENDFOR

\STATE \textcolor{blue}{\textit{// Score leaves, propagate rewards and compute advantage (Eqs.~(5)--(7))}}
\STATE Score all $\ell\in\mathcal{L}$ by Eq.~(\ref{eq:5}); propagate $R$ bottom-up by Eq.~(\ref{eq:6}); compute normalized sibling-relative advantages by Eq.~(\ref{eq:7}).
\STATE \textbf{return} $\mathcal{T}$ and $\mathcal{S}$
\end{algorithmic}
\end{algorithm}

\label{sec:parallel_tree}
\paragraph{Reasoning Tree Representation.}
We model the reasoning process as a tree $\mathcal{T} = (V, E)$, where $V$ is the set of nodes representing partial reasoning states, and $E$ is the set of directed edges connecting them. Each edge corresponds to a continuous reasoning segment from a parent state to a child state. When a \texttt{<look>} marker is detected, a new node is created from the resulting partial reasoning state at that boundary. The reasoning tree begins with the root node, initialized with the input prompt $x_0$ and an empty response $y_0$, whose state is defined as $(x_0, y_0)$.

\paragraph{BFS Parallel Expansion.}
MBPO follows BFS to expand the reasoning tree from the root node, while generating all child branches in parallel at each node. For a node $n$, if its depth satisfies $d_n \ge \mathcal D$, the node is treated as a leaf node, and its reward is computed as:
\begin{equation}
 R(n) = r(x_n, y_n),
\end{equation}
where \( x_n = x_0 \oplus y_1 \oplus \cdots \oplus y_{n-1} \) is the accumulated input sequence from the root to node \( n \), and \( r(x_n, y_n) \) is a score that measures whether the reasoning result \( y_n \) matches the ground truth. If the remaining context length is insufficient to generate valid tokens, node \( n \) is also treated as a leaf node.

Node $n$ is expandable when $d_n < \mathcal D$. We generate $K_{d_n}$ child branches ${z^{(1)}, \ldots, z^{(K_{d_n})}}$ by sampling from $\pi_\theta(\cdot \mid x_n, y_n)$ in one batched forward pass.
\begin{equation}
z^{(i)} \sim \pi_\theta(\cdot \mid x_n, y_n), \quad i = 1, \ldots,  K_{d_n}.
\end{equation}

For each $z^{(i)}$, child node $m$ is instantiated with state:
\begin{equation}
s_{m} = (x_n \oplus y_n, z^{(i)}).
\end{equation}

The policy conditions on the full context $x_n \oplus y_n$, with token-level log-probabilities stored for optimization.

\paragraph{Reward Back-propagation.} After BFS tree construction, MBPO evaluates all leaf nodes in a single batched decoding pass and assigns each leaf node $m$ a scalar reward:
\begin{equation}
R(m) = r(s_m).
\label{eq:5}
\end{equation}
The rewards are then propagated bottom-up through the tree. For each internal node $n$, its reward is defined as the mean reward of its child nodes:
\begin{equation}
R(n) = \frac{1}{|\mathcal{Ch}(n)|} \sum_{m \in \mathcal{Ch}(n)} R(m),
\label{eq:6}
\end{equation}
where $\mathcal{Ch}(n)$ denotes the set of children of node $n$. Based on the propagated rewards, we first compute a local parent-child value difference for each non-root node:
\begin{equation}
A(n) = R(n) - R(\mathrm{parent}(n)).
\label{eq:7}
\end{equation}
This provides an intermediate reward signal for branch evaluation.

\begin{figure*}[bht!]
  \centering
  \includegraphics[width=\textwidth]{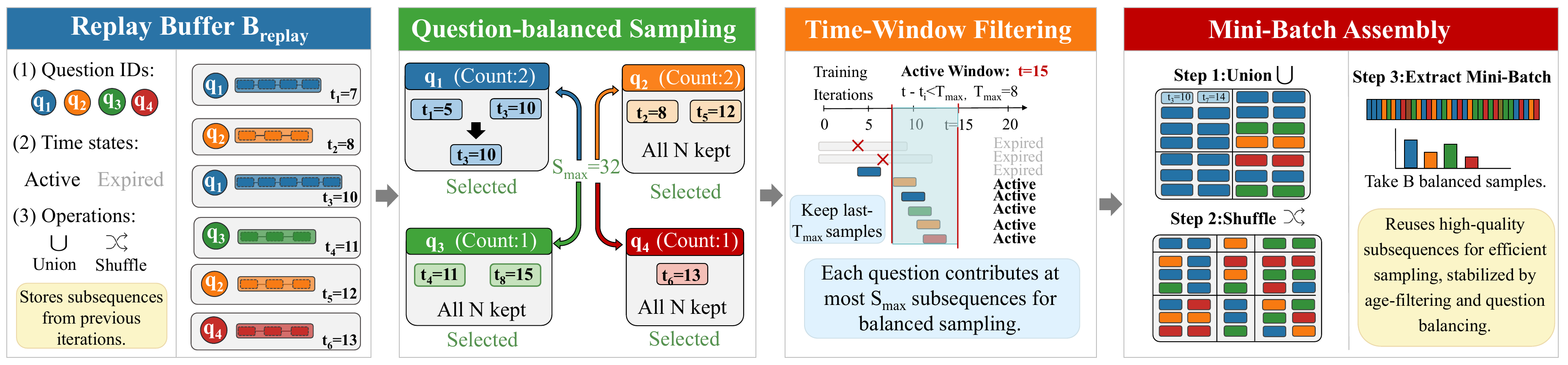}
  \caption{Time-window replay buffer with question-balanced mini-batch sampling, designed to limit stale trajectories and improve sample diversity during training.}
  \label{fig:f4}
  \vspace{-2mm}
\end{figure*}

\subsection{Branch-level Advantage Assignment}
Based on the parallel reasoning tree construction described in Section \ref{sec:parallel_tree}, we represent the sampling process as a tree. Each node $n$ in the tree corresponds to a reasoning subsequence, denoted as ${\mathcal SubSeq}(n)$. This subsequence is generated by extending the sequence of its parent node ${\mathcal Pa}(n)$ with $M$ newly sampled tokens:
\begin{equation}
\begin{aligned}
{\mathcal SubSeq}(n)
  &= \bigl[y^{(n)}_1, \ldots, y^{(n)}_M\bigr], \\
y^{(n)}_t
  &\sim \pi_\theta \left(\cdot \mid
    [\,\mathit{traj}  ({\mathcal Pa}(n)),\, y^{(n)}_{<t}\,]\right).
\end{aligned}
\label{eq:branch_def}
\end{equation}
where $\mathit{traj}({\mathcal Pa}(n))$ denotes the complete reasoning trajectory from the root node to the parent node ${\mathcal Pa}(n)$.

Each node $n$ has a set of child nodes ${\mathcal Ch}(n)$ and a set of sibling nodes ${\mathcal Sib}(n)$. These sibling nodes share the same prompt prefix and sequence length. To enable fair comparison under the same token budget, we estimate the value of the node $n$ in a recursive manner, denoted as $\hat{V}(n)$:
\begin{equation}
\hat{V}(n) =
\begin{cases}
R(x_n, y_n), & \text{if $n$ is a leaf node}, \\
\frac{1}{|{\mathcal Ch}(n)|} \sum_{n' \in {\mathcal Ch}(n)} \hat{V}(n'), & \text{otherwise}.
\end{cases}
\end{equation}
The branch-level advantage of node $n$ is denoted as $A(n)$ and is defined through relative comparison with its sibling nodes. To compare competing continuations under the same parent, we define the branch-level sibling-relative advantage of node $n$ as:
\begin{equation}
A_{\text{sib}}(n)=\frac{\hat{V}(n) - \mathrm{mean}_{n' \in {\mathcal Sib}(n) } \hat{V}(n') }
{\mathrm{std}_{n' \in {\mathcal Sib}(n)} \hat{V}(n') + \epsilon}.
\end{equation}
This normalized sibling-relative advantage is used in PPO.

\subsection{Temporal Replay Buffer}
During training, we generate a large number of reasoning subsequences,
while each parameter update only consumes a limited minibatch. As a result, some high-value subsequences generated earlier may not be fully reused after the model parameters are updated, which can lead to training instability and performance degradation. To address this issue, MBPO introduces a temporal replay buffer that reuses previously generated reasoning subsequences while controlling policy staleness. An illustration of the replay buffer is shown in \Cref{fig:f4}. The replay buffer is defined as:
\begin{equation}
\mathcal{B}_{replay} = \{({\mathcal SubSeq}(i), t_i)\}_{i=1}^{N},
\end{equation}
where $N$ denotes the number of subsequences stored in the buffer, and $t_i$ denotes the training iteration at which subsequence ${\mathcal SubSeq}(i)$ is inserted into the buffer.

\paragraph{Time-Window Filtering.}
At iteration $t$, we restrict the replay buffer to a fixed time window to limit the reuse of outdated subsequences. The active replay set is defined as:
\begin{equation}
\mathcal{B}^{(t)}_{replay} =
\{({\mathcal SubSeq}(i), t_i) \in \mathcal{B}_{\mathrm{replay}} \mid t - t_i < T_{\max}\},
\end{equation}
where $T_{\max}$ is a predefined time horizon, and $t - t_i$ can be interpreted as the age of subsequence ${\mathcal SubSeq}(i)$. This time-window constraint limits the memory length of the replay buffer and avoids reusing outdated subsequences after significant policy changes.

\begin{table*}[t]
\caption{
Performance comparison on representative multimodal reasoning benchmarks. The models marked with $^\ast$ indicate results obtained from re-implementation following the original papers.
To ensure comparability, all RL methods are grouped into a single category and retrained on MMRL18K, with cold-start methods marked by $\dagger$.
The best results among RL-based models are highlighted in \textbf{bold}, and the second-best results are \underline{underlined}.
}
\centering
\fontsize{8pt}{8pt}\selectfont
\renewcommand{\arraystretch}{1.2}

\begin{tabular*}{\textwidth}{@{\extracolsep{\fill}}lccccccc}
\toprule
\textbf{Models} & \textbf{MathVerse} & \textbf{MathVision} & \textbf{MathVista} & \textbf{WeMath} & \textbf{HallBench} & \textbf{ChartQA} & \textbf{Avg.} \\
\midrule

\multicolumn{8}{c}{\textit{Closed-Source Models}} \\
\midrule
GPT-4o \cite{achiam2023gpt}
& 50.8 & 30.4 & 63.8 & 68.8 & 55.0 & --   & - \\
o1 \cite{jaech2024openai}
& 57.0 & 60.3 & 73.9 & --   & --   & --   & - \\
Gemini-2.0 Pro \cite{comanici2025gemini}
& 67.3 & 48.1 & 71.3 & --   & 49.8 & --   & - \\
Claude-3.7-Sonnet \cite{anderson2025comparative}
& 52.0 & 41.3 & 66.8 & 72.6 & 55.4 & --   & - \\

\midrule
\multicolumn{8}{c}{\textit{Open-Source SFT Models}} \\
\midrule
InternVL-2.5-8B \cite{chen2024expanding}
& 39.5 & 17.0 & 64.5 & --   & 50.1 & 79.1 & - \\
InternVL-3-8B \cite{zhu2025internvl3}
& --   & 29.3 & 71.6 & --   & 49.9 & 86.6 & - \\
Qwen2.5-VL-3B$^\ast$ \cite{bai2025qwen2}
& 32.3 & 22.2 & 60.8 & 53.9 & 59.8 & 74.5 & 50.5 \\
Qwen2.5-VL-7B$^\ast$ \cite{bai2025qwen2}
& 43.3 & 22.8 & 67.4 & 68.6 & 59.2 & 78.1 & 56.5 \\

\midrule
\multicolumn{8}{c}{\textit{Reinforcement Learning}} \\
\midrule
R1-VL-7B$^\ast\dagger$ \cite{zhang2025r1}
& 40.5 & 23.8 & 61.6 & 58.7 & 59.2 & 77.5 & 53.5 \\
Vision-R1-7B$^\ast\dagger$ \cite{huang2025vision}
& 46.3 & --   & 69.9 & --   & 59.1 & 81.6 & - \\
R1-OneVision-7B$^\ast\dagger$ \cite{yang2025r1}
& 46.4 & 25.9 & 63.7 & 61.6 & 65.3 & 79.8 & 57.1 \\
OpenVLThinker-7B$^\ast\dagger$ \cite{deng2025openvlthinker}
& 47.7 & 27.0 & 69.7 & 67.5 & 59.4 & 78.8 & 58.4 \\
VLAA-Thinker-7B$^\ast\dagger$ \cite{chen2025sft}
& 48.5 & 26.1 & 69.6 & 67.3 & 67.1 & 81.0 & 59.9 \\
MM-Eureka-Qwen-7B$^\ast$ \cite{meng2025mm}
& 51.6 & 28.1 & 71.5 & 67.4 & 66.7 & 79.0 & 60.7 \\
MMR1-Math-7B$^\ast$ \cite{leng2025mmr1}
& 38.1 & \underline{30.2} & 70.2 & 68.3 & 67.7 & 83.0 & 59.6 \\
ThinkLite-VL-7B$^\ast$ \cite{wang2025sota}
& 45.4 & 28.2 & 71.7 & 69.5 & 69.1 & 82.6 & 61.1 \\
VL-Rethinker-7B$^\ast$ \cite{wang2025vl}
& 50.3 & 28.2 & 70.7 & 69.8 & 68.9 & 80.1 & 61.3 \\
NoisyRollout-7B-K12$^\ast$ \cite{liu2504noisyrollout}
& 49.5 & 27.4 & 71.0 & 70.0 & 68.3 & 82.2 & 61.5 \\
Shuffle-R1-Qwen-7B$^\ast$ \cite{zhu2025shuffle}
& \underline{51.8} & 30.2 & \textbf{74.8} & \underline{70.3} & \underline{69.9} & \underline{84.0} & \underline{63.4} \\
\textbf{MBPO-Qwen-VL-7B}
& \textbf{52.6} & \textbf{30.6} & \underline{74.4} & \textbf{72.6} & \textbf{71.4} & \textbf{87.7} & \textbf{64.9} \\

\bottomrule
\end{tabular*}
\end{table*}

\paragraph{Question-balanced Sampling.}
Based on the active replay set $\mathcal{B}^{(t)}_{\mathrm{replay}}$, we construct a question-balanced mini-batch to reduce data imbalance across questions. For each question $q$, let $S_q$ denote the set of subsequences in $\mathcal{B}^{(t)}_{\mathrm{replay}}$ that are associated with question $q$. To limit the number of subsequences retained for each question, we truncate $S_q$ as:
\begin{equation}
S_q^{\mathrm{trunc}} =
\begin{cases}
{\mathcal RandomSample}(S_q, S_{\max}), & \text{if } |S_q| > S_{\max}, \\
S_q, & \text{otherwise},
\end{cases}
\end{equation}
where $S_{\max}$ denotes the maximum number of subsequences allowed for each question, which is typically set to 32. To construct a mini-batch of size $B$, we combine the truncated sets from all questions and randomly shuffle the result:
\begin{equation}
\mathcal{B}_{\mathrm{batch}} =
\mathrm{Shuffle}\!\left(\bigcup_q S_q^{\mathrm{trunc}}\right)[:B].
\end{equation}

This question-balanced sampling strategy prevents the model from overfitting to a small set of questions that produce many subsequences and improves coverage across training questions.

\section{Experiments}
\subsection{Experimental Setup for MBPO}
\paragraph{Datasets and Benchmarks.}
To evaluate MBPO across data scales, we conduct experiments on multimodal reasoning datasets of different sizes. For small-scale training, we use Geometry3K (Geo3K)~\cite{lu2021inter} and a size-matched subset of MMK12 (K12)~\cite{meng2025mm}. For large-scale training, we construct MMRL-18K using 7.2K filtered samples from MM-Eureka~\cite{meng2025mm} and 11K samples from ThinkLite-VL~\cite{wang2025sota}. We evaluate on the in-domain Geo3K and K12 test sets and six out-of-domain benchmarks: MathVerse~\cite{zhang2024mathverse}, MathVision~\cite{wang2024measuring}, WeMath~\cite{wemath}, MathVista~\cite{mathvista}, HallusionBench~\cite{hallusionbench}, and ChartQA~\cite{chartqa}. These benchmarks cover mathematical reasoning, visual perception, and chart understanding. We use MathRuler and Gemini-2.0-Flash to evaluate free-form and multiple-choice responses, respectively.

\subsection{Main Experimental Results}
\paragraph{Comparison with strong baselines.}
We evaluate on the large-scale MMRL18K dataset. Our MBPO-Qwen-7B model achieves 52.6\% on MathVerse and 30.6\% on MathVision, outperforming the MM-Eureka-Qwen-7B model, which uses only RL, by 1.0\% and 2.5\%, respectively. It also outperforms Vision-R1-7B, and achieves this without cold-start initialization. Additionally, the model scores 87.7\% on ChartQA and 71.4\% on HallBench, which further demonstrates its strong performance in cross-domain reasoning tasks. Moreover, our method shows competitive performance when compared to closed-source models.

\begin{figure}[bht!]
  \centering
  \includegraphics[width=\columnwidth]{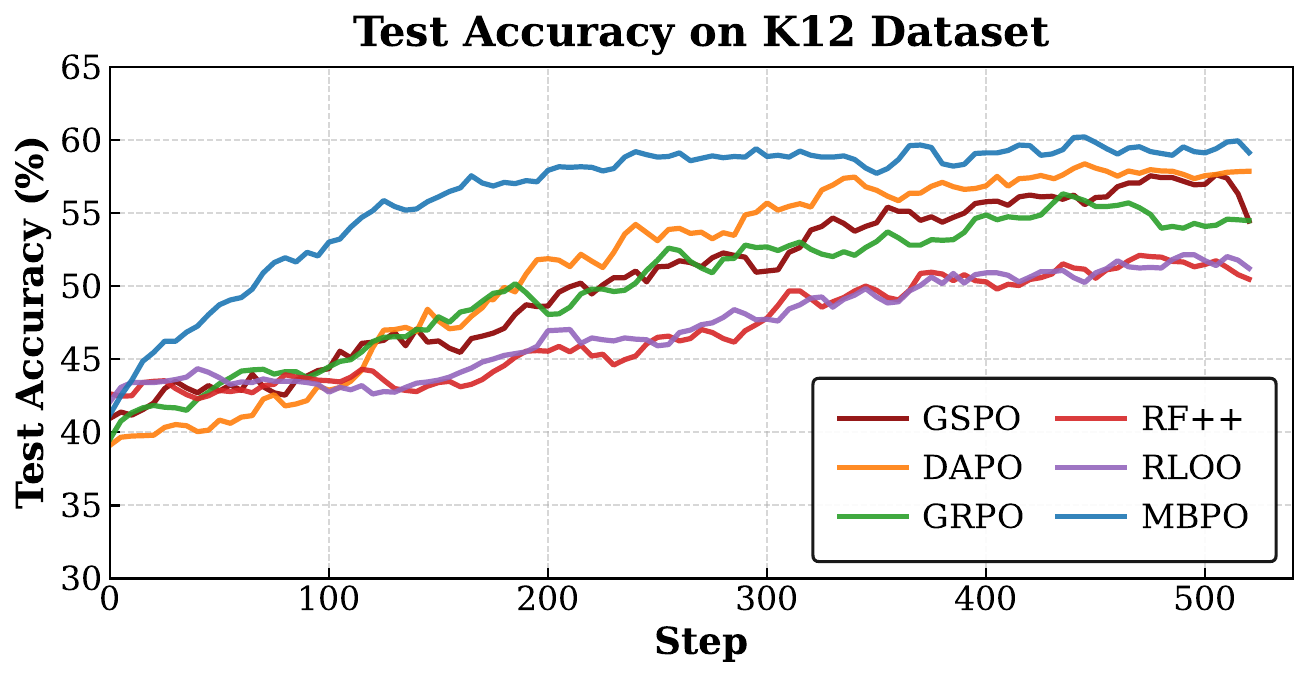}
  \caption{Test accuracy during the training process of MBPO and representative RL algorithms on K12 dataset.}
  \label{fig:f5}
\end{figure}

\begin{table}[bht!]
    \caption{Performance of MBPO on Geo3K dataset compared with GRPO, DAPO and GSPO.}
    \centering
    \fontsize{9pt}{9pt}\selectfont
    \setlength{\tabcolsep}{1mm}
    \begin{tabular}{lcccc}
    \toprule
    \textbf{Method} & \textbf{Geo3K} & \textbf{Math Avg.} & \textbf{HallBench} & \textbf{ChartQA} \\
    \midrule
    Qwen-3B      & 27.83 & 42.35 & 59.83 & 74.54 \\
    + GRPO       & 42.64{\dplus{+14.81}} & 46.46{\dplus{+4.11}} & 61.60{\dplus{+1.77}} & 80.20{\dplus{+5.66}} \\
    + DAPO       & 45.09{\dplus{+17.26}} & 48.57{\dplus{+6.22}} & 62.34{\dplus{+2.51}} & 79.70{\dplus{+5.16}} \\
    + GSPO       & 44.53{\dplus{+16.70}} & 47.68{\dplus{+5.33}} & 63.12{\dplus{+3.29}} & 79.40{\dplus{+4.86}} \\
    \textbf{+ Ours} & \textbf{49.91}{\dplus{+22.08}} & \textbf{49.35}{\dplus{+7.00}} & \textbf{63.60}{\dplus{+3.77}} & \textbf{83.52}{\dplus{+8.98}} \\
    \midrule
    \midrule
    Qwen-7B      & 42.09 & 48.64 & 61.09 & 78.13 \\
    + GRPO       & 51.31{\dplus{+9.22}} & 53.41{\dplus{+4.77}} & 66.31{\dplus{+5.22}} & 79.96{\dplus{+1.83}} \\
    + DAPO       & 53.26{\dplus{+11.17}} & 53.62{\dplus{+4.98}} & 67.10{\dplus{+6.01}} & 81.33{\dplus{+3.20}} \\
    + GSPO       & 52.73{\dplus{+10.64}} & \textbf{53.89}{\dplus{+5.25}} & 67.33{\dplus{+6.24}} & 79.71{\dplus{+1.58}} \\
    \textbf{+ Ours}  & \textbf{56.07}{\dplus{+13.98}} & 53.79{\dplus{+5.15}} & \textbf{68.52}{\dplus{+7.43}} & \textbf{85.15}{\dplus{+7.02}} \\
    \bottomrule
    \end{tabular}
    \label{tab:geo3k-results}
\end{table}

\begin{table}[bht!]
    \caption{Performance of MBPO on K12 dataset compared with GRPO, DAPO and GSPO.}
    \centering
    \fontsize{9pt}{9pt}\selectfont
    \setlength{\tabcolsep}{1mm}
    \begin{tabular}{lcccc}
    \toprule
    \textbf{Method} & \textbf{K12} & \textbf{Math Avg.} & \textbf{HallBench} & \textbf{ChartQA} \\
    \midrule
    Qwen-3B      & 43.10 & 42.35 & 59.83 & 74.54 \\
    + GRPO       & 55.82{\dplus{+12.72}} & 48.04{\dplus{+5.69}} & 63.84{\dplus{+4.01}} & 78.31{\dplus{+3.77}} \\
    + DAPO       & 58.83{\dplus{+15.73}} & 48.24{\dplus{+5.89}} & \textbf{64.88}{\dplus{+5.05}} & 78.42{\dplus{+3.88}} \\
    + GSPO       & 57.63{\dplus{+14.53}} & 48.29{\dplus{+5.94}} & 64.48{\dplus{+4.65}} & 78.20{\dplus{+3.66}} \\
    \textbf{+ Ours} & \textbf{60.14}{\dplus{+17.04}} & \textbf{50.19}{\dplus{+7.84}} & 64.77{\dplus{+4.94}} & \textbf{83.44}{\dplus{+8.90}} \\
    \midrule
    \midrule
    Qwen-7B      & 52.06 & 48.64 & 61.09 & 78.13 \\
    + GRPO       & 65.18{\dplus{+13.12}} & 53.53{\dplus{+4.89}} & 65.80{\dplus{+4.71}} & 82.08{\dplus{+3.95}} \\
    + DAPO       & 67.65{\dplus{+15.59}} & 54.52{\dplus{+5.88}} & 68.71{\dplus{+7.62}} & 81.76{\dplus{+3.63}} \\
    + GSPO       & 67.23{\dplus{+15.17}} & 54.24{\dplus{+5.60}} & 69.02{\dplus{+7.93}} & 81.16{\dplus{+3.03}} \\
    \textbf{+ Ours}  & \textbf{67.68}{\dplus{+15.62}} & \textbf{55.51}{\dplus{+6.87}} & \textbf{69.84}{\dplus{+8.75}} & \textbf{86.36}{\dplus{+8.23}} \\
    \bottomrule
    \end{tabular}
    \label{tab:k12-results}
\end{table}

\paragraph{Performance Comparison with RL Methods.} As shown in Figure~\ref{fig:f5}, MBPO converges faster and achieves higher final accuracy than RLOO, REINFORCE++, GRPO, and DAPO on K12. MBPO also consistently outperforms GRPO, DAPO, and GSPO~\cite{zheng2025groupsequencepolicyoptimization} across all reported metrics on both datasets, as reported in Tables~\ref{tab:geo3k-results} and~\ref{tab:k12-results}. On Geo3K, the Qwen-3B model improves the Math Average by 2.89\% over GRPO and 1.63\% over GSPO, while achieving a 2.00\% gain on HallusionBench. The Qwen-7B model exhibits similarly consistent improvements. These results demonstrate the effectiveness of MBPO across datasets and model scales, particularly on challenging multimodal reasoning tasks.

\subsection{Why Does MBPO Work?}
\begin{figure}[bht!]
  \centering

  \begin{subfigure}{\columnwidth}
    \centering
    \includegraphics[width=\linewidth]{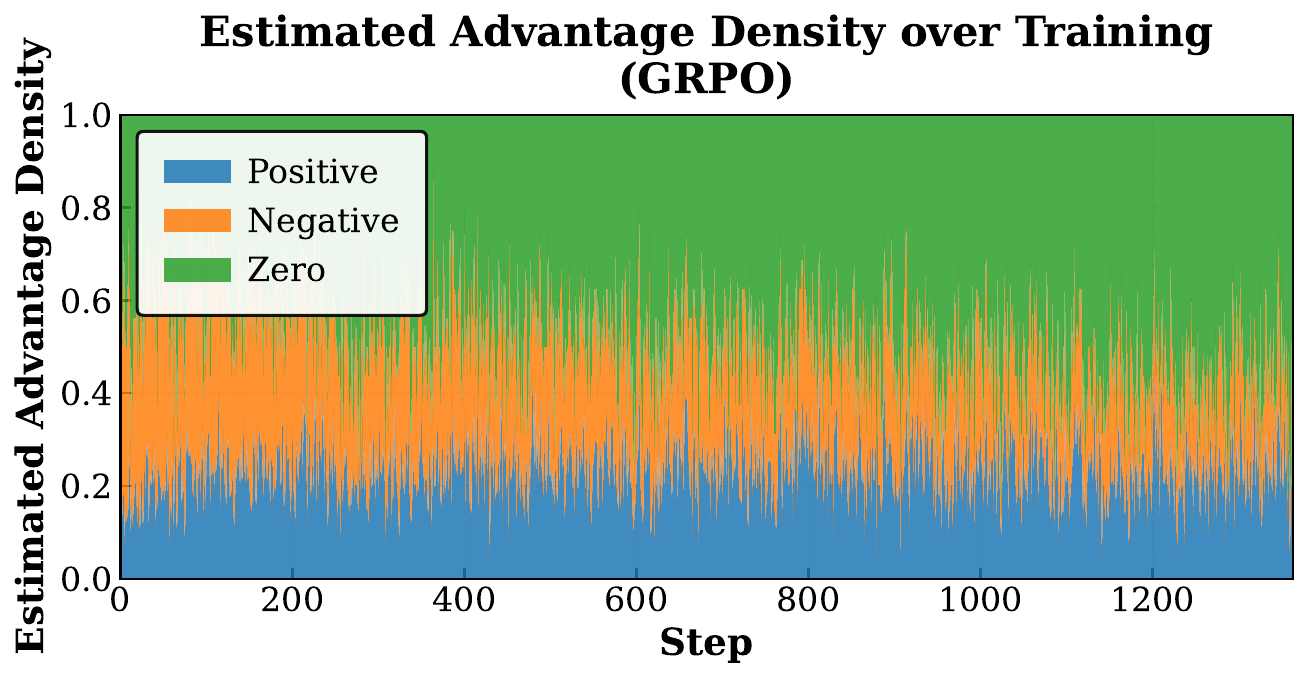}
    \label{fig:6a}
  \end{subfigure}

  \begin{subfigure}{\columnwidth}
    \centering
    \includegraphics[width=\linewidth]{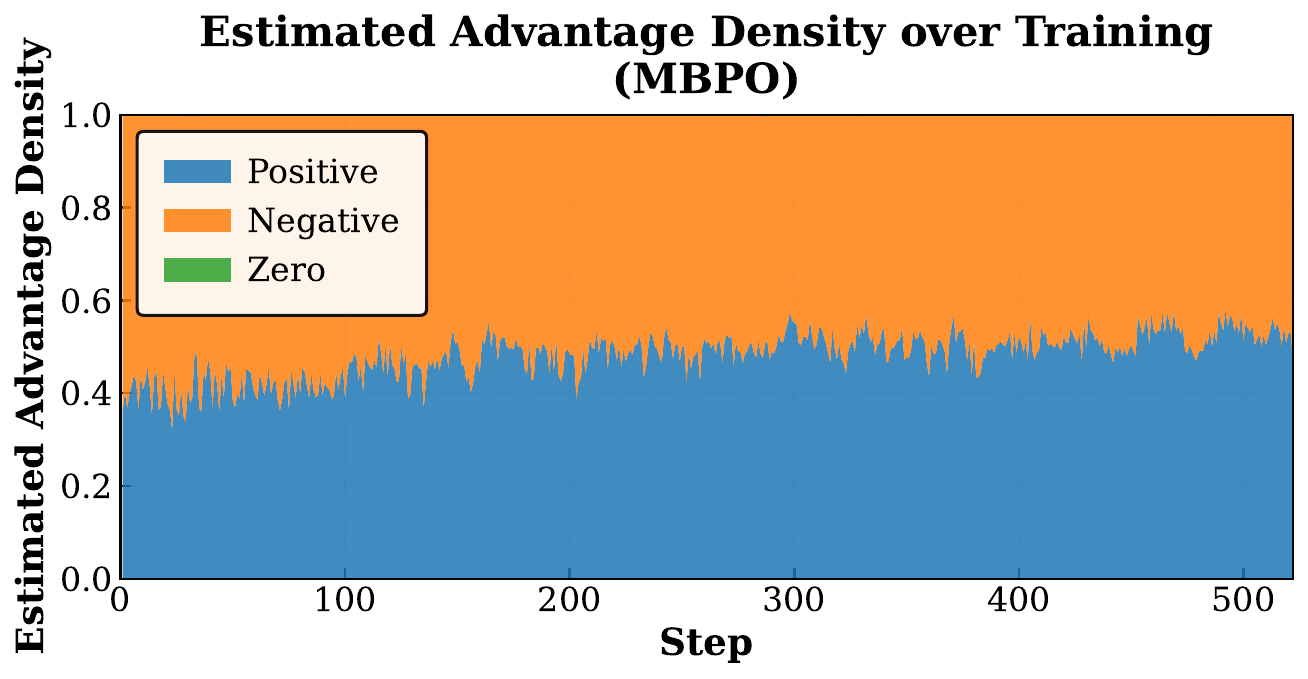}
    \label{fig:f6b}
  \end{subfigure}

  \caption{Estimated advantage distribution across training steps for GRPO and MBPO on Geo3K dataset.}
  \label{fig:f6}
\end{figure}

\paragraph{Advantage Density Analysis.}
To analyze the evolution of the advantage distribution during training, we estimate the per-step advantage density by decomposing it into positive, negative and zero components, as shown in Figure \ref{fig:f6}. With GRPO, the distribution is largely dominated by zero-advantage values throughout training, while the positive and negative components exhibit substantial fluctuations. This indicates that many sampled updates provide limited learning signal and result in a sparse advantage distribution. In contrast, MBPO shows a different profile, with little mass at zero and density concentrated on non-zero advantages. As training progresses, the proportion of positive advantages steadily increases, accompanied by a gradual decline in negative ones. These results suggest that MBPO promotes a more informative and consistently positive advantage distribution during optimization.

\begin{figure}[bht!]
  \centering
  \includegraphics[width=\columnwidth]{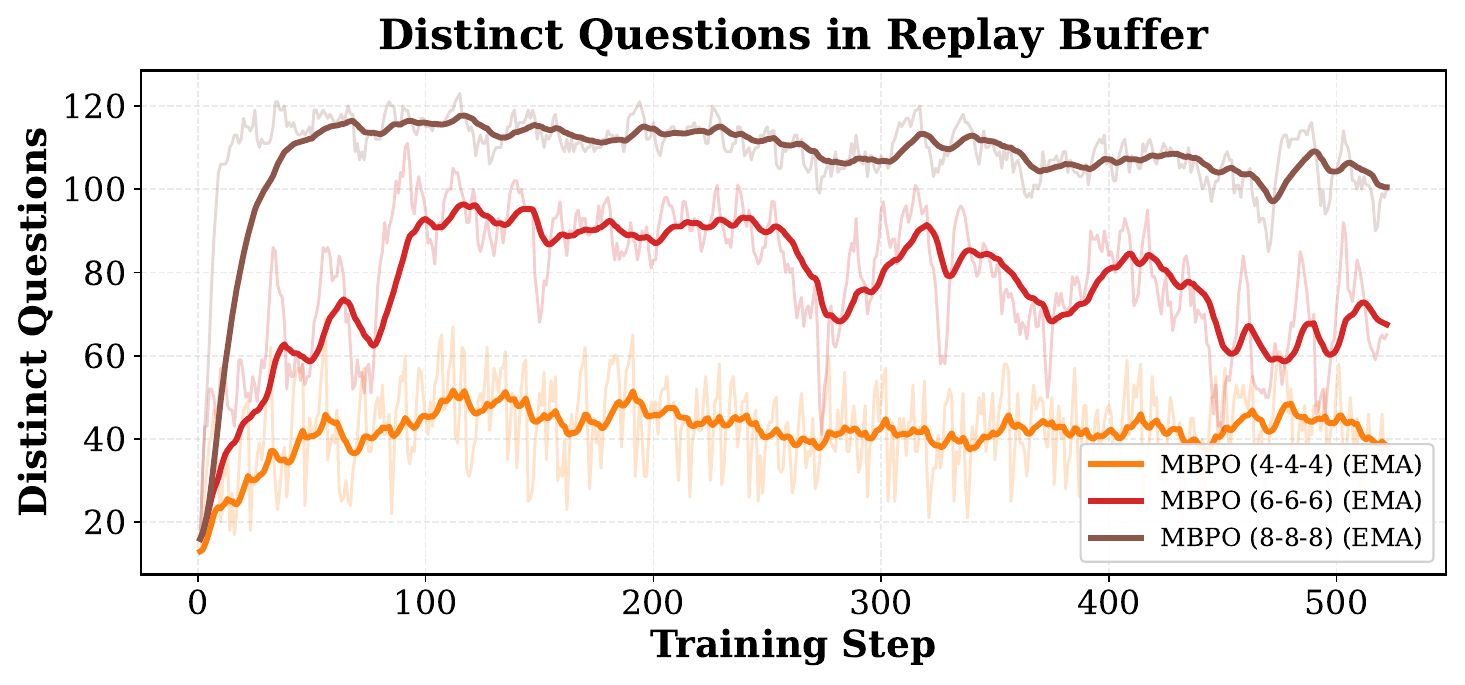}
  \caption{Replay buffer diversity under different MBPO tree configurations,
measured by the number of distinct questions stored in the replay buffer.}
  \label{fig:f8}
\end{figure}

\paragraph{Diversity in the Replay Buffer.}
To better understand the role of the replay buffer in MBPO, we compute statistics on the types of questions in the buffer during training and analyze the changes in sample diversity by comparing different tree structure configurations (such as 4-4-4, 6-6-6, and 8-8-8). As shown in Figure \ref{fig:f8}, MBPO (8-8-8) rapidly increases question diversity in the early stages and maintains a high level of diversity throughout training, with only a slight decrease in the final stage. In contrast, MBPO (6-6-6) reaches a moderate plateau but fluctuates more, with several sharp drops in diversity. MBPO (4-4-4) saturates at a lower diversity level and remains roughly flat, indicating lower exploration and a narrower replay buffer. These results suggest that larger tree structures help support broader and more stable replay diversity, potentially providing richer learning signals for the optimization process.

\begin{figure*}[bht!]
  \begin{center}

    \begin{minipage}[t]{0.32\textwidth}
      \centering
      \includegraphics[width=\linewidth]{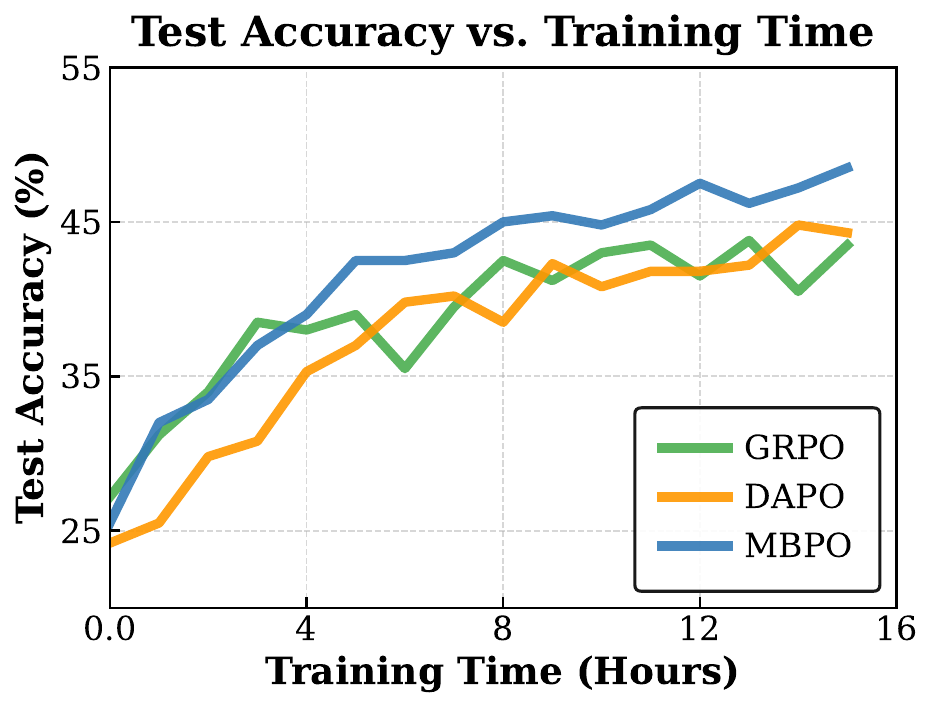}
      \subcaption*{(a)}
      \label{fig:f7c}
    \end{minipage}\hfill
    \begin{minipage}[t]{0.32\textwidth}
      \centering
      \includegraphics[width=\linewidth]{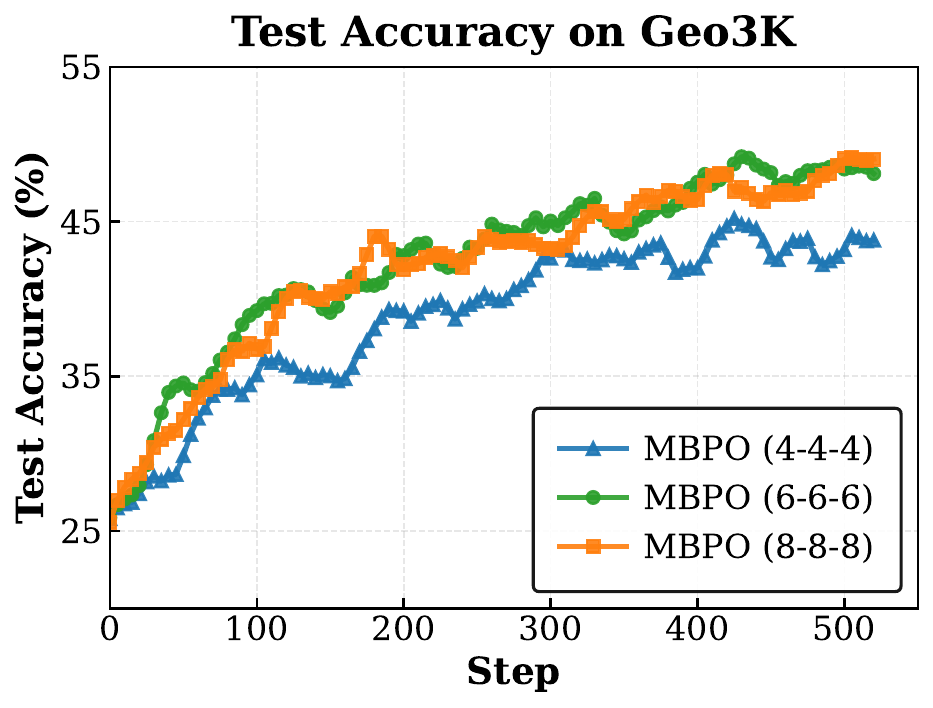}
      \subcaption*{(b)}
      \label{fig:f7a}
    \end{minipage}\hfill
    \begin{minipage}[t]{0.32\textwidth}
      \centering
      \includegraphics[width=\linewidth]{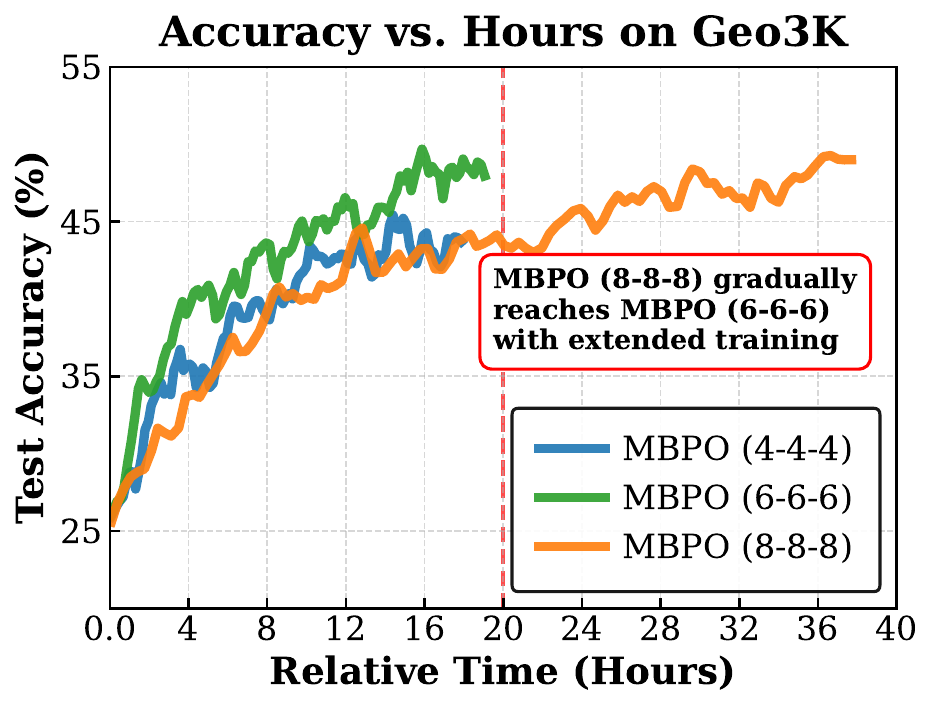}
      \subcaption*{(c)}
      \label{fig:f7b}
    \end{minipage}\hfill

    \caption{
        (a) Test accuracy over training time (hours) for MBPO, GRPO, and DAPO on Geo3K.
        (b) Test accuracy over training steps for different tree structures.
        (c) Test accuracy over training time (hours) for different tree structures.
    }
    \label{fig:ablation_study}
  \end{center}
  \vspace{-2mm}
\end{figure*}

\paragraph{Self-Correction Behavior} For each generated response, we analyze the text following the \texttt{<Look>} marker and identify self-correction steps via explicit revision cues at the token level (e.g., \texttt{however}, \texttt{but}, \texttt{actually}, \texttt{wait}, \texttt{reconsider}) and phrase level (e.g., \texttt{on second thought}, \texttt{that is incorrect}), and report the rate at which such revisions lead to correct final answers. As shown in Tables~\ref{tab:mbpo_behavior_main} and \ref{tab:grpo_dapo_mbpo_sc}, MBPO exhibits increasingly frequent and earlier self-corrections as training progresses, with the successful correction rate rising from 4.7\% to 29.8\%, whereas GRPO, DAPO, and GSPO show limited self-correction and rarely perform multi-step revisions. These results confirm that MBPO provides clear learning signals that encourage the model to revisit visual inputs and perform corrections at critical reasoning nodes.

\begin{table}[bht!]
    \caption{\label{tab:mbpo_behavior_main}
Self-correction statistics during MBPO training. Response length is measured in tokens from the first token after \texttt{<think>}. SC: self-correction rate; Succ. Rate: successful correction rate; Median 1st Corr. Pos.: median token position of the first correction.}
    \centering
    \fontsize{9pt}{9pt}\selectfont
    \setlength{\tabcolsep}{0.6mm}
    \begin{tabular}{lcccc}
    \toprule
    \begin{tabular}[c]{@{}c@{}}\textbf{Training}\\\textbf{Steps}\end{tabular} &
    \begin{tabular}[c]{@{}c@{}}\textbf{Median 1st}\\\textbf{Corr.\ Pos.} $\downarrow$\end{tabular} &
    \begin{tabular}[c]{@{}c@{}}\textbf{Avg. Response}\\\textbf{Length (tokens)} $\uparrow$\end{tabular} &
    \begin{tabular}[c]{@{}c@{}}\textbf{SC Rate}\\\textbf{(\%)} $\uparrow$\end{tabular} &
    \begin{tabular}[c]{@{}c@{}}\textbf{Succ. Rate}\\\textbf{(\%)} $\uparrow$\end{tabular} \\
    \midrule
    0   & 129.0 & 170.5 & 13.3 & 4.7  \\
    250 & 115.0 & 236.3 & 45.3 & 20.5 \\
    524 & 108.5 & 239.2 & 60.9 & 29.8 \\
    \bottomrule
    \end{tabular}
\end{table}

\begin{table}[bht!]
    \caption{Comparison of self-correction behaviors among GRPO, DAPO, MBPO, and GSPO.}
    \centering
    \fontsize{9pt}{9pt}\selectfont
    \setlength{\tabcolsep}{1mm}
    \begin{tabular}{@{}lcccc@{}}
    \toprule
    \textbf{Method} &
    \begin{tabular}[c]{@{}c@{}}\textbf{Avg. Response}\\\textbf{Length (tokens)} $\uparrow$\end{tabular} &
    \begin{tabular}[c]{@{}c@{}}\textbf{SC Rate}\\\textbf{(\%)} $\uparrow$\end{tabular} &
    \begin{tabular}[c]{@{}c@{}}\textbf{$\ge$2C Rate}\\\textbf{(\%)} $\uparrow$\end{tabular} &
    \begin{tabular}[c]{@{}c@{}}\textbf{Succ. Rate}\\\textbf{(\%)} $\uparrow$\end{tabular} \\
    \midrule
    GRPO & 196.2 & 20.1 & 7.7  & 7.8 \\
    DAPO & 198.5 & 12.3 & 3.8  & 7.4 \\
    GSPO & 243.8 & 29.9 & 16.21 & 19.3 \\
    MBPO & 239.2 & 60.9 & 44.6 & 29.8 \\
    \bottomrule
    \end{tabular}
    \label{tab:grpo_dapo_mbpo_sc}
\end{table}

\paragraph{Compute-Controlled Analysis}
A key concern for tree-based policy optimization is whether the observed gains simply reflect increased rollout expenditure. To address this, we compare MBPO with GRPO under matched cumulative compute budgets, keeping the number of training segments per update fixed at 1,024 for both methods. As shown in Table~\ref{tab:compute_matched_mbpo_grpo} and Figure~\ref{fig:ablation_study}(a), MBPO achieves higher accuracy at every aligned compute checkpoint and under the same wall-clock training time, with improvements ranging from +2.0 to +8.7 points. These results indicate that the gains stem from more effective use of computation rather than a larger rollout budget. Although tree construction introduces additional cost per optimization step, the BFS batching strategy groups all same-depth nodes into a single vLLM call with \(n=K\) parallel sampling, keeping the per-epoch GPU cost comparable: 19.48 GPU-hours for MBPO versus 19.32 for GRPO on 4 GPUs (less than 1\% overhead). MBPO thus maintains a favorable accuracy--time trade-off and consistently outperforms GRPO and DAPO throughout training.

\subsection{Ablation Study}
\paragraph{Ablation on Tree Structure} We compare three tree structure configurations: MBPO (4-4-4), MBPO (6-6-6), and MBPO (8-8-8). As shown in Figure~\ref{fig:ablation_study}(b), larger tree structures yield better performance in later training stages, with MBPO (6-6-6) and MBPO (8-8-8) both outperforming MBPO (4-4-4). Figure~\ref{fig:ablation_study}(c) shows that MBPO (8-8-8) gradually approaches the accuracy of MBPO (6-6-6) given extended training time, indicating that MBPO (6-6-6) offers a better efficiency--accuracy trade-off. 

\begin{table}[t]
    \caption{Compute-matched performance comparison between GRPO and MBPO on Geo3K. Each row fixes the cumulative training TFLOPs and reports the optimization steps and test accuracy achieved by each method under the same compute budget. Bold numbers indicate the best result.}
    \centering
    \fontsize{8.5pt}{8.5pt}\selectfont
    \setlength{\tabcolsep}{3.5mm}
    \begin{tabular}{@{}cccccc@{}}
    \toprule
    \begin{tabular}[c]{@{}c@{}}\textbf{Cumul.}\\\textbf{TFLOPs}\end{tabular} &
    \begin{tabular}[c]{@{}c@{}}\textbf{GRPO}\\\textbf{Steps}\end{tabular} &
    \begin{tabular}[c]{@{}c@{}}\textbf{GRPO}\\\textbf{Acc}\end{tabular} &
    \begin{tabular}[c]{@{}c@{}}\textbf{MBPO}\\\textbf{Steps}\end{tabular} &
    \begin{tabular}[c]{@{}c@{}}\textbf{MBPO}\\\textbf{Acc}\end{tabular} &
    \textbf{$\Delta$} \\
    \midrule
    0.469M & 50  & 29.8 & 50  & \textbf{33.8} & +4.0 \\
    0.917M & 100 & 32.6 & 95  & \textbf{40.6} & +8.0 \\
    1.387M & 150 & 37.8 & 140 & \textbf{39.8} & +2.0 \\
    1.892M & 200 & 39.9 & 190 & \textbf{41.9} & +2.0 \\
    2.354M & 250 & 37.9 & 235 & \textbf{42.4} & +4.5 \\
    2.832M & 300 & 39.8 & 280 & \textbf{45.3} & +5.5 \\
    3.309M & 350 & 40.3 & 325 & \textbf{45.6} & +5.3 \\
    3.788M & 400 & 37.1 & 370 & \textbf{45.8} & +8.7 \\
    4.242M & 450 & 41.4 & 415 & \textbf{48.4} & +7.0 \\
    4.688M & 500 & 44.1 & 455 & \textbf{47.1} & +3.0 \\
    4.871M & 520 & 43.4 & 475 & \textbf{48.1} & +4.7 \\
    \bottomrule
    \end{tabular}
    \label{tab:compute_matched_mbpo_grpo}
    \vspace{-2mm}
\end{table}

\begin{table}[t]
\centering
\fontsize{9pt}{10.5pt}\selectfont
\setlength{\tabcolsep}{2.6mm}
\renewcommand{\arraystretch}{1.03}
\caption{Ablation of branching strategies and segment granularity on Geo3K. Bold numbers indicate the best result.}
\label{tab:branching_segment_ablation}
\begin{tabular}{l l c c}
\toprule
\textbf{Setting} &
\textbf{Branching} &
\makecell[c]{\textbf{Geo3K} \\\textbf{Acc.}} &
\makecell[c]{\textbf{Avg. Seg.} \\\textbf{Len.}} \\
\midrule
GRPO & \textit{flat} & 42.64 & -- \\
\midrule
MBPO &\textit{ hybrid ($M$=100 + \texttt{<look>})} & \textbf{49.91} & 83.6 \\
\hdashline
& \textit{$M$ only (fixed $M$=100)} & 48.35 & 100.0 \\
& \textit{\texttt{<look>} only} & 45.17 & 287.4 \\
& \textit{hybrid ($M$=50 + \texttt{<look>})} & 46.82 & 43.2 \\
& \textit{hybrid ($M$=200 + \texttt{<look>})} & 48.53 & 163.5 \\
\bottomrule
\end{tabular}
\end{table}

\paragraph{Effect of Branching Strategy and Segment Length}
\label{sec:branch_seg}
As shown in Table~\ref{tab:branching_segment_ablation}, all MBPO variants outperform GRPO (42.64), which confirms the effectiveness of tree-structured exploration. Fixed-$M$ branching achieves 48.35 with an average segment length of 100.0 tokens, offering stable and frequent splits, but it may divide the reasoning path at positions that are not well aligned with semantic boundaries. Visual branching via \texttt{<look>} obtains 45.17, but produces much longer segments (avg.\ 287.4 tokens), suggesting that \texttt{<look>} is triggered too infrequently and therefore leads to shallow trees with limited branching. The hybrid strategy combines both mechanisms, reducing the average segment length to 83.6 tokens while achieving the best accuracy of 49.91, which indicates that \texttt{<look>} helps refine segment boundaries while fixed-$M$ maintains sufficient branching density. For segment length, $M=50$ over-fragments the reasoning process, resulting in shorter segments (avg.\ 43.2 tokens) but lower accuracy (46.82), whereas $M=200$ produces overly long segments (avg.\ 163.5 tokens, 48.53) that restrict branching diversity. By comparison, $M=100$ provides the best balance between branching frequency and segment coherence.

\section{Conclusion}
In this work, we propose Multi-Branch Policy Optimization (MBPO), a reinforcement learning framework for MLLMs that improves credit assignment through branch-level advantage estimation on a reasoning tree. With the temporal replay buffer, MBPO reuses useful segments and makes learning more stable for long reasoning tasks. Experiments on multiple multimodal reasoning datasets demonstrate the effectiveness and training stability of MBPO.

\section{Limitations and Future Work}
However, MBPO has limitations, including high training costs, challenges in adapting to broader multimodal settings, and difficulty in large-scale transfer. A key future direction is to predict whether a partial reasoning path is worth continuing, so we can stop low-value branches early and focus on more promising ones. We also plan to build stronger datasets and benchmarks, and to scale MBPO to longer contexts and larger models.

\clearpage
\begin{acks}
This work was supported by the National Natural Science Foundation of China under Grants U25B6003, U25A20534, 62472295, and 62406036; the National Social Science Fund of China under Grant 24BGL131; the National Key Research and Development Program of China under Grant 2024YFC3308500; the Beijing Municipal Natural Science Foundation under Grant L251042; the China Postdoctoral Science Foundation under Grant 2025M781457; the Key Laboratory of Knowledge Mining and Services for Integrated Publishing in Education under Grant KT20250804; the State Key Laboratory of Networking and Switching Technology under Grant NST20250110; the Chengdu Science and Technology Bureau Project under Grant 2025-YF05-00391-SN; the Sichuan Science and Technology Planning Project under Grant 2024NSFSC0521; and the Luzhou City School-Local-Enterprise-Academy Science and Technology Cooperation Project under Grant 2024XDY200.
\end{acks}

\bibliographystyle{ACM-Reference-Format}
\bibliography{mbpo}

\end{document}


\maketitle

\appendix

\section{Theoretical Analysis}
\label{sec:theory_appendix}

\noindent\textbf{Proposition 1. Effective branch signals are dominated by critical decision points.}

Consider one policy update with the node set $\mathcal{N}$ and $|\mathcal{N}|=N$.
We define the unnormalized sibling advantage as
\[
\tilde A(n)=\hat V(n)-\operatorname{mean}_{n'\in\mathrm{Sib}(n)}\hat V(n').
\]
We define the effective-signal ratio as
\[
\mathrm{VAR}_{\text{branch}}
=\frac{1}{N}\sum_{n\in\mathcal{N}}\mathbb{I}\left(|\tilde A(n)|>\varepsilon\right).
\]
We assume that each parent is either non-critical or critical.
A non-critical parent means that all its children have the same true value.
A critical parent means that there exist two children whose true values differ by at least $\Delta>0$.
We also assume a bounded estimation error for all nodes in the update:
\[
|\hat V(n)-v(n)|\le \delta.
\]
We further assume $\Delta>8\delta$ so that the threshold window below is not empty.
If the threshold satisfies
\begin{equation}
\label{eq:threshold_window_merged_appendix}
2\delta<\varepsilon<\frac{\Delta}{2}-2\delta,
\end{equation}
then all effective terms in $\mathrm{VAR}_{\text{branch}}$ come from branches under critical parents.
Moreover, let $M$ denote the number of critical parents whose full sibling groups are included in the update batch.
Then we have
\begin{equation}
\label{eq:var_lower_bound_merged_appendix}
\mathrm{VAR}_{\text{branch}}\ge \frac{M}{N}.
\end{equation}
Therefore, under the same sampling and update budget, effective learning signals concentrate on critical decision points, while branches under non-critical parents are pushed below the threshold.

\paragraph{Proof.}
Fix a parent node $p$ with children $\mathrm{Ch}(p)=\{n_1,\dots,n_K\}$.

\paragraph{Step 1: Branches under non-critical parents contribute no effective terms.}
If $p$ is non-critical, then $v(n_1)=\cdots=v(n_K)=c$ for some constant $c$.
Let $\overline{\hat V}$ be the mean estimated value of the children:
\[
\overline{\hat V}=\frac{1}{K}\sum_{j=1}^K\hat V(n_j).
\]
For any child $n_i$, we have
\[
\tilde A(n_i)=\hat V(n_i)-\overline{\hat V}
=\big(\hat V(n_i)-c\big)-\big(\overline{\hat V}-c\big).
\]
By the triangle inequality,
\[
|\tilde A(n_i)|
\le |\hat V(n_i)-c|+|\overline{\hat V}-c|.
\]
Using the error bound for all children, we have $|\hat V(n_i)-c|\le\delta$ and
\[
|\overline{\hat V}-c|
=\left|\frac{1}{K}\sum_{j=1}^K(\hat V(n_j)-c)\right|
\le \frac{1}{K}\sum_{j=1}^K|\hat V(n_j)-c|
\le \delta.
\]
Therefore, for all children under a non-critical parent,
\[
|\tilde A(n_i)|\le 2\delta.
\]
Since $\varepsilon>2\delta$ by Eq.~\eqref{eq:threshold_window_merged_appendix}, we get
\[
\mathbb{I}(|\tilde A(n_i)|>\varepsilon)=0.
\]

\paragraph{Step 2: Each critical parent yields at least one effective branch.}
If $p$ is critical, then there exist two children $n_i,n_j$ such that $|v(n_i)-v(n_j)|\ge \Delta$.
Let $\bar v$ be the mean true value of the children, and let $\overline{\hat V}$ be the mean estimated value:
\[
\bar v=\frac{1}{K}\sum_{m=1}^K v(n_m),
\qquad
\overline{\hat V}=\frac{1}{K}\sum_{m=1}^K \hat V(n_m).
\]
Let $v_{\max}=\max_m v(n_m)$ and $v_{\min}=\min_m v(n_m)$.
Then $v_{\max}-v_{\min}\ge \Delta$ and $\bar v\in[v_{\min},v_{\max}]$.
This implies
\[
\max\{|v_{\max}-\bar v|,\ |\bar v-v_{\min}|\}\ge \frac{v_{\max}-v_{\min}}{2}\ge \frac{\Delta}{2}.
\]
So there exists a child $n_k$ such that
\[
|v(n_k)-\bar v|\ge \frac{\Delta}{2}.
\]
For this child, we have
\[
\tilde A(n_k)=\hat V(n_k)-\overline{\hat V}
=\big(v(n_k)-\bar v\big)+\big(\hat V(n_k)-v(n_k)\big)-\big(\overline{\hat V}-\bar v\big).
\]
By the reverse triangle inequality,
\[
|\tilde A(n_k)|
\ge |v(n_k)-\bar v|-|\hat V(n_k)-v(n_k)|-|\overline{\hat V}-\bar v|.
\]
Using the error bound, we have $|\hat V(n_k)-v(n_k)|\le \delta$ and
\[
|\overline{\hat V}-\bar v|
=\left|\frac{1}{K}\sum_{m=1}^K(\hat V(n_m)-v(n_m))\right|
\le \frac{1}{K}\sum_{m=1}^K|\hat V(n_m)-v(n_m)|
\le \delta.
\]
Therefore,
\[
|\tilde A(n_k)|\ge \frac{\Delta}{2}-2\delta.
\]
Since $\varepsilon<\frac{\Delta}{2}-2\delta$ by Eq.~\eqref{eq:threshold_window_merged_appendix}, we have
\[
|\tilde A(n_k)|>\varepsilon,
\]
so its indicator equals one.

\paragraph{Step 3: Conclude dominance and the lower bound.}
By Step 1, all branches under non-critical parents have indicator zero.
By Step 2, each such critical parent contributes at least one child branch with indicator one.
Let $M$ be the number of critical parents whose full sibling groups are included in the update batch.
Then the total number of ones in the sum is at least $M$, which gives
\[
\mathrm{VAR}_{\text{branch}}
=\frac{1}{N}\sum_{n\in\mathcal{N}}\mathbb{I}\left(|\tilde A(n)|>\varepsilon\right)
\ge \frac{M}{N}.
\]
This also proves that all effective terms come from branches under critical parents.
\hfill $\square$

\paragraph{Implication.}
Under the same sampling and update budget, $\mathrm{VAR}_{\text{branch}}$ is mainly contributed by branches under critical decision points and admits a positive lower bound.
At the same time, $|\tilde A|$ on non-critical branches is pushed below the threshold, so effective branch-level signals focus on decision points that directly affect the final reward.
In contrast, trajectory-level methods often spread the same advantage over many irrelevant tokens or segments, which dilutes the learning signal and makes credit assignment less accurate, especially in long reasoning trajectories.

\section{Implementation and Training Details}
\label{sec:impl_train_appendix}

\begin{table}[bht!]
\centering
\caption{Hyperparameter settings used in our experiments.}
\label{tab:hyperparameters_appendix}
\small
\setlength{\tabcolsep}{10pt}
\begin{tabular}{ll}
\toprule
\textbf{Hyperparameter} & \textbf{Value} \\
\midrule
Max pixels & 1,000,000 \\
Min pixels & 262,144 \\
Max prompt length & 2048 (1024 for Qwen2.5-VL-3B) \\
Max response length & 2048 (1024 for Qwen2.5-VL-3B) \\
Target train batch size & 32 (16 for Qwen2.5-VL-3B) \\
Num episodes per iteration & 1024 \\
Max samples per question & 32 \\
Rollout size & 16 \\
Max tokens per node & 100 \\
Branching factors & $[6,6,6]$ \\
Train policy temperature & 1.0 \\
Train policy top-$p$ & 1.0 \\
Train context size & 2048 (1024 for Qwen2.5-VL-3B) \\
Initial KL coef & 0.01 \\
Learning rate & $1\times10^{-6}$ \\
Epochs per iteration & 1 \\
Prob mask threshold & 0.9 \\
Evaluation context size & 2048 \\
Evaluation temperature & 0 \\
Evaluation top-$p$ & 1.0 \\
Evaluation samples & 1 \\
\bottomrule
\end{tabular}
\end{table}

\subsection{Implementation Details}
We implement MBPO based on the verl framework \cite{Sheng_2025} by extending it with tree-based trajectory sampling and the temporal replay buffer.
All experiments are conducted using the Qwen2.5-VL-3B/7B-Instruct models.
We use a batch size of 16 and a learning rate of $1\times 10^{-6}$.
PPO optimization is performed with AdamW and KL regularization, with the KL coefficient set to $0.01$.
MBPO adopts a three-level tree structure, with the branching factor set to 6 at each level and the maximum number of tokens per node limited to 100.
During training, the temporal replay buffer samples up to 1024 segments per iteration, with at most 32 per question and an age limit of 8.

\subsection{Branch Segmentation Strategy}
\label{sec:branch_seg}
Each expansion generates up to $M=100$ tokens per branch.
After generation, MBPO checks whether a \texttt{<look>} tag appears within the generated segment.
If detected at position $p$, the segment is truncated at $p$ and a new branching node is created, aligning the tree structure with a point where the model re-examines visual information.
If no \texttt{<look>} tag is found, the full $M$-token segment is used as the branch.
This hybrid strategy encourages branching at semantically meaningful vision-language decision points when such boundaries arise naturally, while the fixed-length fallback ensures stable cross-branch comparisons when the model does not explicitly revisit visual content within the token budget.

\subsection{Evaluation Details}
We use vLLM \cite{kwon2023efficientmemorymanagementlarge} as the inference engine and adopt VLMEvalKit \cite{duan2025vlmevalkitopensourcetoolkitevaluating} for standardized evaluation.
Accuracy is reported for each dataset.
Following prior work \cite{liu2504noisyrollout, zhu2025shuffle}, we use Gemini-2.0-Flash \cite{team2023gemini} to first extract the predicted answer from model outputs and then compare it with the ground truth.
The reproduced results in the main paper are obtained using the same evaluation pipeline and settings.

\subsection{Hyperparameters and Training Resources}
We conduct all experiments based on the verl training framework.
Table~\ref{tab:hyperparameters_appendix} lists the critical hyperparameters used in our experiments.
For settings not included in the table, we keep the default configurations of verl.
For compute resources, we train Qwen2.5-VL-3B-Instruct with 4 NVIDIA A800 40GB GPUs, and we train Qwen2.5-VL-7B-Instruct with 8 NVIDIA A800 40GB GPUs.
In the main paper, we use the same training and evaluation setup for GRPO, DAPO, and GSPO as in our method.

\section{Dataset Curation and Prompt Templates}
\label{sec:data_prompt_appendix}

\subsection{Dataset Curation}
To evaluate model performance at a larger data scale, we curate a dataset named MMRL18K by combining two complementary sources with a unified filtering and merging procedure.
We first randomly sample 8K examples from the K12 dataset released in MM-Eureka and denote this subset as K12-8K.
We then re-filter K12-8K using Qwen2.5-VL-7B-Instruct as a critic model.
Concretely, we follow the data curation prompt template shown in Figure~\ref{fig:data_filter_prompt} to generate responses and check whether each sample can be solved in a reliable way.
Samples that are consistently answered correctly by the critic model are treated as easy cases, since they provide limited training signal in our reinforcement learning setting.
Based on this criterion, we remove the 0.8K easiest samples from K12-8K, which increases the overall difficulty and reduces the number of trivial instances in the training data.

\begin{figure*}[t]
\centering
\includegraphics[width=\textwidth]{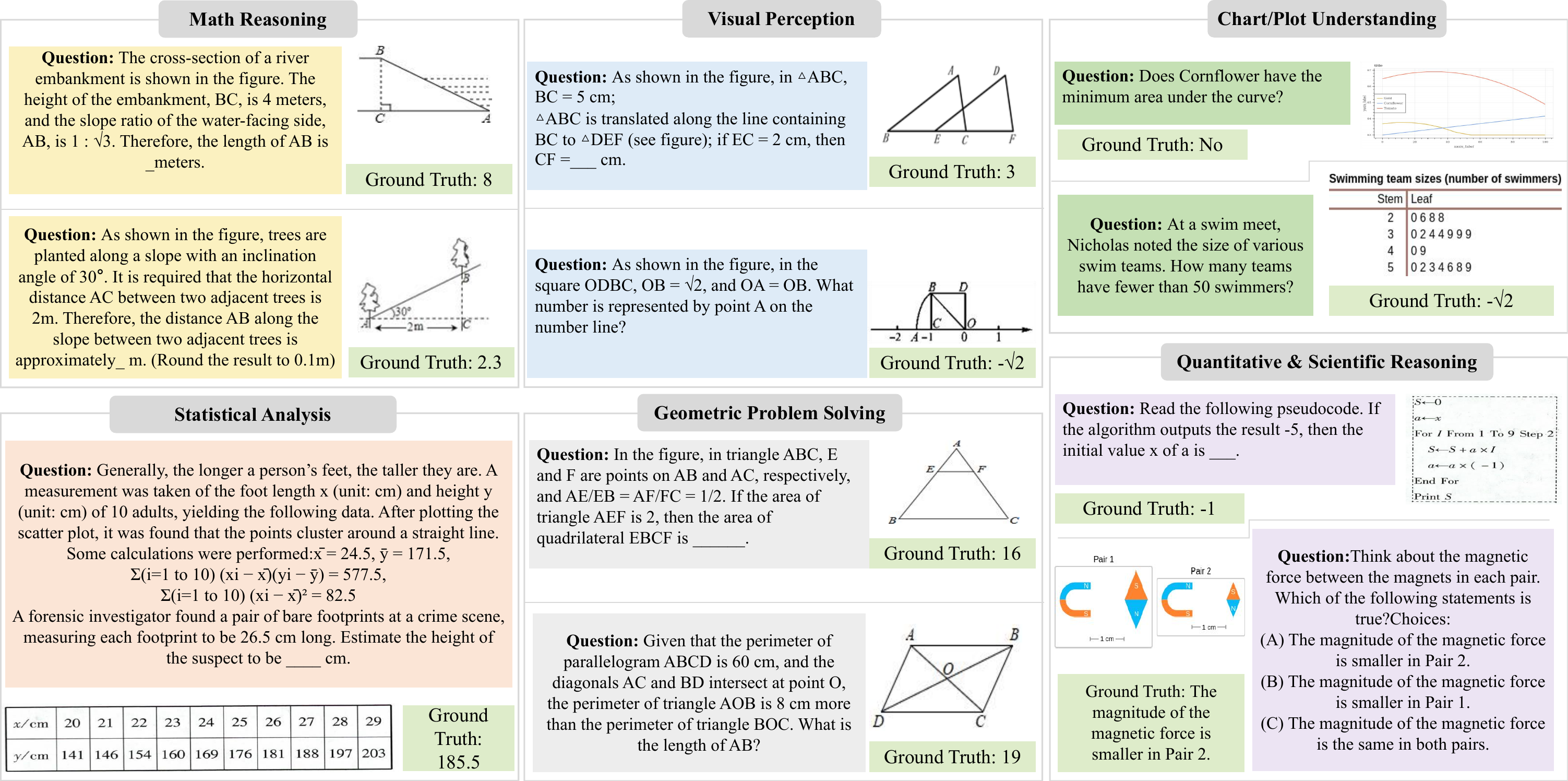}
\caption{Examples from MMRL18K, covering diverse multimodal reasoning categories, including math reasoning, visual perception, chart and plot understanding, statistical analysis, geometric problem solving, and quantitative scientific reasoning.}
\label{fig:f9_appendix}
\end{figure*}

After re-filtering, we merge the remaining K12 data with an additional 11K dataset from ThinkLite-VL.
This dataset contains harder and more diverse multimodal reasoning examples.
After merging, we obtain MMRL18K with a total size of 18K samples.
Throughout this process, we keep a consistent input and output format, and we apply the same preprocessing rules and validation checks to both sources.
Overall, this curation strategy produces a larger and more challenging training set, while avoiding an over-representation of easy questions that can reduce the effectiveness of reinforcement learning updates.
Example samples from MMRL18K are shown in Figure~\ref{fig:f9_appendix}.

\subsection{Prompt for Data Filtering}
We randomly sample 8K K12-style problems from MM-Eureka as our candidate pool.
To remove samples that are too easy and provide limited learning signals for RL, we use Qwen2.5-VL-7B-Instruct as a critic model.
For each sample, the model first solves the problem and then outputs a KEEP or DROP decision based on simple rules, such as direct reading or one-step arithmetic.
We remove 0.8K easy samples and keep about 7.2K samples.
The prompt used for this filtering procedure is shown in Figure~\ref{fig:data_filter_prompt}.

\subsection{Prompt Template for RL}
We use a unified prompt template for RL training so that model outputs are consistent across tasks and datasets.
The system prompt instructs the model to first reason within \texttt{<think>...</think>} tags, include one or more \texttt{<look>...</look>} blocks where it pays special attention to certain visual information in the figure, and provide the final answer in \texttt{\textbackslash boxed\{\}}.
Concretely, the system prompt is

\smallskip
\noindent\fbox{%
\parbox{0.97\columnwidth}{%
\small\texttt{You FIRST think about the reasoning process as an internal monologue and then provide the final answer. The reasoning process MUST BE enclosed within <think> </think> tags. You must include one or more <look>...</look> blocks enclosing the parts of the reasoning where you pay special attention to certain visual information in the figure. The final answer MUST BE put in \textbackslash boxed\{\}.}%
}}
\smallskip

\noindent The \texttt{<look>} tag serves as a vision-language boundary marker: when it appears within a generated segment during tree construction, MBPO truncates the segment at that position and creates a branching node (see Section~\ref{sec:branch_seg}).
This design encourages the model to produce explicit visual re-examination points, which MBPO exploits as semantically meaningful branching boundaries.

\section{Additional Experimental Results}
\label{sec:more_exp_appendix}

\subsection{Detailed Experimental Results}

\begin{table*}[t]
\centering
\caption{Detailed performance on out-of-domain math reasoning benchmarks for models trained on Geo3K and K12.}
\label{tab:mv_mvsta_wemath_avg_appendix}
\small
\setlength{\tabcolsep}{12pt}
\begin{tabular}{lccccc}
\toprule
\textbf{Method} & \textbf{MathVerse} & \textbf{MathVision} & \textbf{MathVista} & \textbf{WeMath} & \textbf{Avg.} \\
\midrule
Qwen2.5-VL-3B & 32.38 & 22.21 & 60.87 & 53.94 & 42.35 \\
\quad +Ours (Geo3K) & 44.59 (+12.21) & 25.00 (+2.79) & 64.00 (+3.13) & 63.79 (+9.85) & 49.35 (+7.00) \\
\quad +Ours (K12) & 45.03 (+12.65) & 25.32 (+3.11) & 65.10 (+4.23) & 65.32 (+11.38) & 50.19 (+7.84) \\
\midrule
Qwen2.5-VL-7B & 43.32 & 22.81 & 68.65 & 59.78 & 48.64 \\
\quad +Ours (Geo3K) & 47.62 (+4.30) & 26.74 (+3.93) & 72.32 (+3.67) & 68.48 (+8.70) & 53.79 (+5.15) \\
\quad +Ours (K12) & 49.88 (+6.56) & 27.58 (+4.77) & 73.67 (+5.02) & 70.91 (+11.13) & 55.51 (+6.87) \\
\bottomrule
\end{tabular}
\end{table*}

\begin{table*}[t]
\centering
\caption{Detailed performance on out-of-domain math reasoning benchmarks for models trained on math-lvl3to5-8k.}
\label{tab:llm_results_appendix}
\small
\setlength{\tabcolsep}{8pt}
\begin{tabular}{lcccccc}
\toprule
\textbf{Method} & \textbf{math-lvl3to5-8k} & \textbf{AIME24} & \textbf{MATH500} & \textbf{AMC} & \textbf{OlympiadBench} & \textbf{Minerva-MATH} \\
\midrule
Qwen2.5-Math-1.5B & -- & -- & -- & -- & -- & -- \\
\quad +GRPO & 71.4 & 18.53 & 73.00 & 45.78 & 32.24 & 18.75 \\
\quad +Ours & 75.0 & 17.67 & 74.20 & 46.89 & 34.76 & 22.80 \\
\midrule
DeepSeek-R1-Distill-Qwen-1.5B & -- & -- & -- & -- & -- & -- \\
\quad +GRPO & 78.40 & 15.76 & 76.20 & 49.39 & 34.47 & 23.89 \\
\quad +Ours & 82.80 & 16.66 & 77.60 & 54.22 & 35.36 & 24.27 \\
\bottomrule
\end{tabular}
\end{table*}

Table~\ref{tab:mv_mvsta_wemath_avg_appendix} reports detailed results on out-of-domain math reasoning benchmarks for models trained on Geo3K and K12.
In addition to the average score, we present performance on each benchmark, including MathVerse, MathVision, MathVista, and WeMath.
The results show consistent gains for both the 3B and 7B models after training on Geo3K or K12, and the improvements are especially clear on MathVerse and WeMath.
These detailed scores help explain the changes in the average results reported in the main paper.

\subsection{Additional Ablation Study}
\label{sec:additional_ablation_study_appendix}
We conduct the ablation study using the Qwen2.5-VL-3B-Instruct model, as shown in Table~\ref{tab:ablation_study_appendix}.
The results show that mean aggregation for reward back-propagation performs best, giving the highest accuracy on both Geo3K and K12.
Removing the temporal replay buffer reduces performance, highlighting its importance in reusing high-value samples during training.
Similarly, removing the branch-level advantage leads to lower accuracy, showing that finer credit assignment is necessary for effective learning.
Among the reward back-propagation methods, softmax performs better than max, confirming that a smoother, more balanced reward aggregation improves the learning process.
These results further support our design choice: combining temporal replay with branch-level advantage assignment leads to more effective optimization.

\begin{table}[t]
\centering
\caption{Ablation study on Geo3K and K12 using Qwen2.5-VL-3B-Instruct.}
\label{tab:ablation_study_appendix}
\small
\setlength{\tabcolsep}{3pt}
\renewcommand{\arraystretch}{1.05}
\begin{tabularx}{\columnwidth}{@{}>{\raggedright\arraybackslash}X c c c@{}}
\toprule
\textbf{Setting} &
\makecell[c]{\textbf{Reward}\\\textbf{Aggregation}} &
\makecell[c]{\textbf{Geo3K}\\\textbf{Acc.}} &
\makecell[c]{\textbf{K12}\\\textbf{Acc.}} \\
\midrule
MBPO & mean & 49.91 & 60.14 \\
\midrule
\multicolumn{4}{@{}l@{}}{\textbf{Component ablations}} \\
w/o Replay Buffer & mean & 45.83 (-4.08) & 53.91 (-6.23) \\
\makecell[l]{w/o Branch-level Adv.\\(traj-level)} & mean & 44.76 (-5.15) & 52.28 (-7.86) \\
\midrule
\multicolumn{4}{@{}l@{}}{\textbf{Reward back-propagation variants}} \\
Variant 1 & max & 47.53 (-2.38) & 55.53 (-4.61) \\
Variant 2 & softmax & 48.69 (-1.22) & 58.47 (-1.67) \\
\bottomrule
\end{tabularx}
\end{table}

\subsection{Generalization to LLMs}
\label{sec:generalization_llm}
\begin{figure}[t]
\centering
\begin{minipage}[t]{0.32\textwidth}
  \centering
  \includegraphics[width=\linewidth]{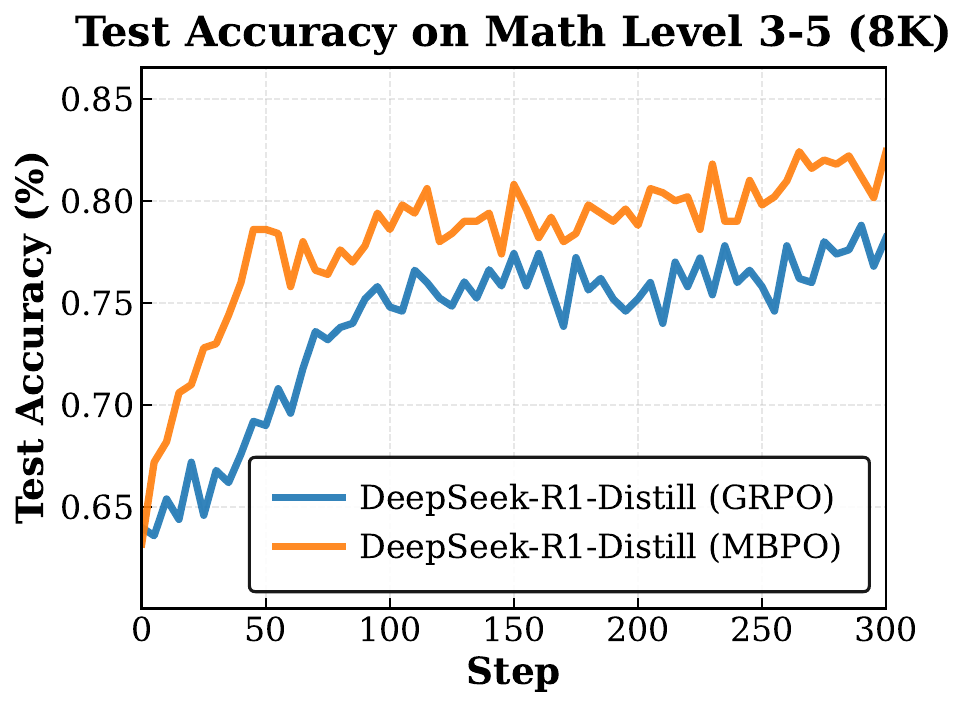}
  \par\vspace{2pt}
  {\small (a)}
\end{minipage}\hfill
\begin{minipage}[t]{0.32\textwidth}
  \centering
  \includegraphics[width=\linewidth]{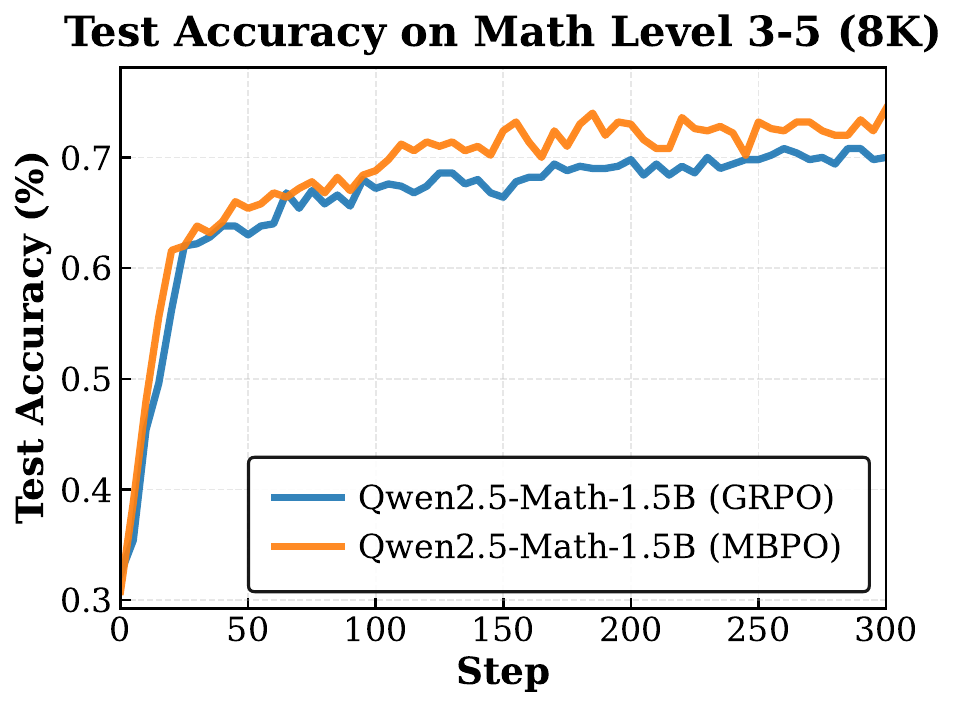}
  \par\vspace{2pt}
  {\small (b)}
\end{minipage}\hfill
\caption{(a) Test accuracy on Math Level 3 to 5 (8K) across training steps for DeepSeek-R1-Distill-Qwen-1.5B. (b) Test accuracy on Math Level 3 to 5 (8K) across training steps for Qwen2.5-Math-1.5B.}
\label{fig:f10_appendix}
\end{figure}

To test whether our method also works for text-only models, we run experiments on two math LLMs, Qwen2.5-Math-1.5B and DeepSeek-R1-Distill-Qwen-1.5B, trained on the math-lvl3to5-8k \cite{liu2025understandingr1zeroliketrainingcritical} dataset.
Table~\ref{tab:llm_results_appendix} reports the out-of-domain results on six math reasoning benchmarks, including AIME24 \cite{li2024numinamath}, MATH500 \cite{lightman2023let}, AMC \cite{li2024numinamath}, OlympiadBench \cite{he2024olympiadbench}, and Minerva-MATH \cite{lewkowycz2022solving}.
Compared with GRPO, our method achieves better performance on most benchmarks for both models.
Figures~\ref{fig:f10_appendix}(a) and \ref{fig:f10_appendix}(b) show the learning curves during training and make the accuracy differences across steps clearer for both models.
These results indicate that our method can generalize from MLLMs to LLMs and can improve out-of-domain math reasoning.

\begin{table}[t]
\centering
\caption{Results on representative REC splits with Qwen2.5-VL-3B. MBPO consistently outperforms both the base model and GRPO.}
\label{tab:rec_results_appendix}
\small
\renewcommand{\arraystretch}{1.05}
\begin{tabular*}{\columnwidth}{@{\extracolsep{\fill}}lccc@{}}
\toprule
\textbf{Method} & \textbf{RefCOCO B} & \textbf{RefCOCO+ B} & \textbf{RefCOCOg} \\
\midrule
Base & 75.64 & 66.93 & 72.39 \\
GRPO & 77.3 & 67.6 & 76.4 \\
MBPO & \textbf{79.2} & \textbf{71.5} & \textbf{80.5} \\
\bottomrule
\end{tabular*}

\vspace{0.35em}
\begin{minipage}{\columnwidth}
\footnotesize
\textit{Note.} The base model may output bounding boxes in different formats, including absolute, relative (0--1), and normalized (0--1000) coordinates. For IoU evaluation, relative boxes are converted to absolute coordinates. This may lead to slightly lower accuracy than the official technical report on some test sets.
\end{minipage}
\end{table}

\subsection{Generalization to REC}
\label{sec:generalization_rec}
To evaluate whether MBPO can generalize beyond multimodal reasoning benchmarks, we further consider Referring Expression Comprehension (REC), a vision-language grounding task that requires predicting the target bounding box from a natural-language expression. We train Qwen2.5-VL-3B on a 20K subset and report results on three representative and relatively challenging splits: RefCOCO testB, RefCOCO+ testB, and RefCOCOg test.

As shown in Table~\ref{tab:rec_results_appendix}, MBPO consistently outperforms both the base model and GRPO on all three reported splits, demonstrating the effectiveness of branch-level policy optimization beyond math reasoning. Notably, these splits are also relatively challenging: RefCOCO testB and RefCOCO+ testB involve more object-centric scenes that require finer visual discrimination, while RefCOCOg test contains longer and more compositional referring expressions. In such settings, accurate grounding depends on revisiting critical local evidence before making the final prediction, which is closely related to the role of the \texttt{<look>} mechanism in MBPO. These results suggest that MBPO is not only effective for mathematical reasoning, but also transfers well to multimodal tasks grounded in basic visual scenes.

\section{Case Studies}
\label{sec:case_appendix}
In this section, we present three representative case studies covering math reasoning, statistical analysis, and visual perception. Tables~\ref{tab:branch_adv_example_appendix}--\ref{tab:case2_oblique_projection_appendix} show that MBPO is more reliable than both the base model and GRPO across different problem types. In the math reasoning example, the base model and GRPO rely on incorrect geometric assumptions, while MBPO identifies the correct geometric constraint and derives the right equation. In the statistical analysis example, the base model and GRPO incorrectly substitute the regression equation directly, whereas MBPO distinguishes the fitted value from the observed entry and correctly infers $t=3$. In the visual perception example, the base model and GRPO fail to identify the correct projected shape, while MBPO revisits the key visual evidence and selects the correct option.

Notably, the \texttt{<look>} blocks in MBPO outputs indicate points where the model explicitly re-examines critical visual or structured evidence. The highlighted \texttt{<look>} blocks mark semantically meaningful decision points that can trigger visual-uncertainty-driven branch expansion during training, thereby more effectively shaping subsequent reasoning behavior.

\begin{figure}[t]
\centering
\begin{tcolorbox}[
  colback=gray!10,
  colframe=gray!60,
  coltitle=white,
  colbacktitle=gray!55,
  title=Data Curation Prompt Template,
  fonttitle=\bfseries,
  boxrule=0.6pt,
  arc=2pt,
  left=8pt,
  right=8pt,
  top=6pt,
  bottom=6pt
]
\small\ttfamily
You are a strict data filter for building a reinforcement learning training set for multimodal math reasoning.

You will receive: \\
- \textless Image\textgreater \\
- A problem text

\textbf{Your tasks:} \\
1) Solve the problem using the image and output the final answer. \\
2) Decide whether the sample is TOO EASY for RL training and should be dropped.

\textbf{Decision rules:} \\
- DROP only if the sample is clearly EASY and provides limited RL value. \\
- TOO EASY means the answer can be obtained by direct reading from the image, or by one short step, or by very basic arithmetic. \\
- KEEP if the problem needs two or more meaningful steps, geometry reasoning, algebraic manipulation, combining multiple visual cues, or careful interpretation of a figure or chart. \\
- If you are unsure, choose KEEP. \\
- Prefer KEEP in borderline cases.

\textbf{Output format:} \\
- Output only valid JSON, with no extra text. \\
- Use the exact keys and allowed values below. \\
- Keep ``reason'' under 20 words.

JSON schema: \\
\{ \\
\quad ``final\_answer'': ``\textless answer\textgreater'', \\
\quad ``estimated\_steps'': 1 or 2 or 3, \\
\quad ``decision'': ``KEEP'' or ``DROP'', \\
\quad ``difficulty'': ``EASY'' or ``MEDIUM'' or ``HARD'', \\
\quad ``confidence'': ``LOW'' or ``MEDIUM'' or ``HIGH'', \\
\quad ``reason'': ``\textless short reason\textgreater'' \\
\}

Now process this sample: \\
\textless Image\textgreater \\
Problem: \{problem\_text\} \\
Output:
\end{tcolorbox}
\caption{Prompt template used for data filtering.}
\label{fig:data_filter_prompt}
\end{figure}

\begin{table*}[t]
\centering
\caption{A case study on a math reasoning problem.}
\label{tab:branch_adv_example_appendix}
\small
\setlength{\tabcolsep}{4pt}
\renewcommand{\arraystretch}{1.12}
\begin{tabularx}{\textwidth}{@{}l>{\raggedright\arraybackslash}Xcc@{}}
\toprule
\textbf{Method} & \textbf{Reasoning Process} & \textbf{Final Answer} & \textbf{Correct?} \\
\midrule
\multicolumn{4}{@{}p{\textwidth}@{}}{
\begin{minipage}[t]{0.16\textwidth}
\vspace{0pt}
\centering
\includegraphics[width=\linewidth]{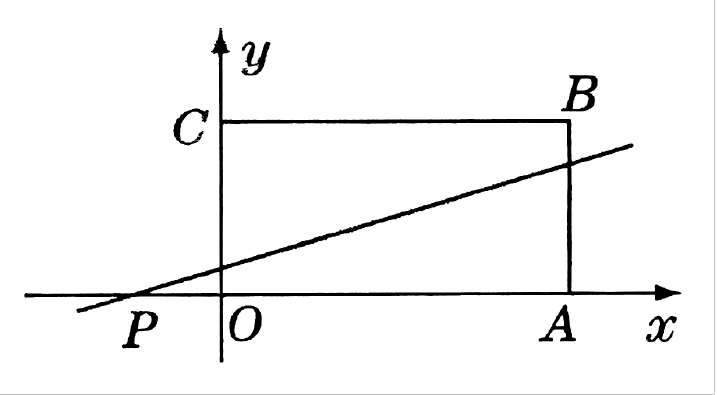}
\end{minipage}\hfill
\begin{minipage}[t]{0.81\textwidth}
\vspace{0pt}
\textbf{Question:} As shown in the figure, rectangle $OABC$ is in the Cartesian coordinate system, with $OA=4$ and $OC=2$. A line passing through point $P(-1,0)$ divides the area of rectangle $OABC$ into two equal parts. The equation of this line is \_.\\
\textbf{Ground Truth:} $y=\dfrac{1}{3}x+\dfrac{1}{3}$.
\end{minipage}
} \\
\midrule

Base Model
&
\begin{minipage}[t]{\linewidth}
\ttfamily
<think>

To find the equation of the line passing through point $P(-1,0)$ that divides the area of rectangle $OABC$ into two equal parts, we proceed as follows:

1) Determine the area of rectangle $OABC$: $OA=4$, $OC=2$, so the area is $4\times 2=8$, and each part has area $4$.

2) Let the line be $y=mx+b$. Since it passes through $P(-1,0)$, $0=m(-1)+b$, so $b=m$, and the line is $y=mx+m$.

3) Determine $m$ by setting the area above the line to be $4$:
\[
\text{Area}
=\int_{-1}^{4}(mx+m)\,dx
=\left[\frac{mx^2}{2}+mx\right]_{-1}^{4}
=\frac{25m}{2}.
\]

</think>

Thus,
\[
\frac{25m}{2}=4 \Rightarrow m=\boxed{\frac{8}{25}}.
\]

\end{minipage}
& $\boxed{y=\frac{8}{25}x+\frac{8}{25}}$ & No \\
\midrule

GRPO
&
\begin{minipage}[t]{\linewidth}
\ttfamily
<think>

To find the equation of the line passing through $P(-1,0)$ that divides the area of rectangle $OABC$ into two equal parts, we follow these steps:

1) The rectangle has area $4\times 2=8$, so each part has area $4$.

2) Let the line be $y=mx+c$. Since it passes through $P(-1,0)$, $c=m$, so the line is $y=mx+m$.

3) The line must pass through the origin of the rectangle and the point $(4,2)$, so
\[
m=\frac{2-0}{4-0}=\frac{1}{2}.
\]

</think>
\end{minipage}
& $\boxed{y=\frac{1}{2}x+\frac{1}{2}}$ & No \\
\midrule

MBPO
&
\begin{minipage}[t]{\linewidth}
\ttfamily
<think>

To find the equation of the equal-area bisecting line, I should identify the geometric point that such a line must pass through.

\lookstart

\looktext{First inspect the rectangle dimensions in the figure: $OA=4$ and $OC=2$, so the center of the rectangle should be the midpoint of these side lengths.}

\lookend

The area of the rectangle is $OA\times OC=4\times 2=8$, so each part has area $4$.

A line that bisects the area of a rectangle must pass through its center, whose coordinates are
\[
\left(\frac{0+4}{2},\frac{0+2}{2}\right)=(2,1).
\]

\lookstart

\looktext{Now verify the two points that determine the line: the given point is $P(-1,0)$, and the geometric constraint gives the center $(2,1)$.}

\lookend

Therefore, the line passes through $P(-1,0)$ and $(2,1)$, so its slope is
\[
m=\frac{1-0}{2-(-1)}=\frac{1}{3}.
\]

Using $y=mx+b$ and point $(-1,0)$ gives
\[
0=\frac{1}{3}(-1)+b \Rightarrow b=\frac{1}{3}.
\]

</think>

Hence, the equation is
$\boxed{y=\frac{1}{3}x+\frac{1}{3}}$

\end{minipage}
& $\boxed{y=\frac{1}{3}x+\frac{1}{3}}$ & Yes \\
\bottomrule
\end{tabularx}
\end{table*}

\begin{table*}[t]
\centering
\caption{A case study on a statistical analysis problem.}
\label{tab:case3_linear_regression_appendix}
\small
\setlength{\tabcolsep}{4pt}
\renewcommand{\arraystretch}{1.12}
\begin{tabularx}{\textwidth}{@{}l>{\raggedright\arraybackslash}Xcc@{}}
\toprule
\textbf{Method} & \textbf{Reasoning Process} & \textbf{Final Answer} & \textbf{Correct?} \\
\midrule
\multicolumn{4}{@{}p{\textwidth}@{}}{
\begin{minipage}[t]{0.16\textwidth}
\vspace{0pt}
\centering
\includegraphics[width=\linewidth]{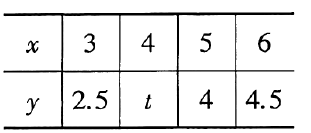}
\end{minipage}\hfill
\begin{minipage}[t]{0.81\textwidth}
\vspace{0pt}
\textbf{Question:} A factory recorded the production of product $A$ (in tons) and the corresponding energy consumption $y$ (in tons) after implementing energy-saving and consumption-reducing technological upgrades, as shown in the table. Based on the data provided in the table, the linear regression equation of $y$ with respect to $x$ is $\hat{y}=0.7x+0.35$. What is the value of $t$ in the table?\\
\textbf{Ground Truth:} $3$.
\end{minipage}
} \\
\midrule

Base Model
&
\begin{minipage}[t]{\linewidth}
\ttfamily
<think>

To find $t$, we start by using the given linear regression equation $\hat{y}=0.7x+0.35$.

Substitute $x=4$ into the equation:
\[
\hat{y}=0.7\cdot 4+0.35=3.15
\]

</think>

Thus, $t$=$\boxed{3.15}$.

\end{minipage}
& $\boxed{3.15}$ & No \\
\midrule

GRPO
&
\begin{minipage}[t]{\linewidth}
\ttfamily
<think>

To find $t$, we start with the given regression equation $\hat{y}=0.7x+0.35$.

The table provides the following data points:
\[
\begin{array}{c|c}
x & y \\
\hline
3 & 2.5 \\
4 & t \\
5 & 4 \\
6 & 4.5 \\
\end{array}
\]

We substitute $x=4$ into the regression equation:
\[
\hat{y}=0.7\cdot 4+0.35=3.15
\]
</think>

Thus, the value of $t$ is $\boxed{3.15}$.

\end{minipage}
& $\boxed{3.15}$ & No \\
\midrule

MBPO
&
\begin{minipage}[t]{\linewidth}
\ttfamily
<think>

To find $t$, I should distinguish the regression prediction from the actual observed table entry.

\lookstart

\looktext{First read the table carefully: the value at $x=4$ is an unknown observation $t$, not the fitted value directly given by the regression equation.}

\lookend

The data points are
\[
\begin{array}{c|c}
x & y \\
\hline
3 & 2.5 \\
4 & t \\
5 & 4 \\
6 & 4.5 \\
\end{array}
\]

\lookstart

\looktext{Now use the key regression property that the fitted line passes through $(\bar{x},\bar{y})$, which links the unknown table entry to the sample mean.}

\lookend

First, calculate the mean of the $x$-values:
\[
\bar{x}=\frac{3+4+5+6}{4}=4.5.
\]

Then, the mean of the $y$-values is
\[
\bar{y}=\frac{2.5+t+4+4.5}{4}=\frac{11+t}{4}.
\]

Since the regression line passes through $(\bar{x},\bar{y})$, substitute $\bar{x}=4.5$ and $\bar{y}=\frac{11+t}{4}$ into the regression equation:
\[
\frac{11+t}{4}=0.7\cdot 4.5+0.35=3.5.
\]
</think>

Hence,
\[
11+t=14 \quad \Rightarrow \quad t=\boxed{3}.
\]

\end{minipage}
& $\boxed{3}$ & Yes \\
\bottomrule
\end{tabularx}
\end{table*}

\begin{table*}[t]
\centering
\caption{A case study on a visual perception problem.}
\label{tab:case2_oblique_projection_appendix}
\small
\setlength{\tabcolsep}{4pt}
\renewcommand{\arraystretch}{1.12}
\begin{tabularx}{\textwidth}{@{}l>{\raggedright\arraybackslash}Xcc@{}}
\toprule
\textbf{Method} & \textbf{Reasoning Process} & \textbf{Final Answer} & \textbf{Correct?} \\
\midrule
\multicolumn{4}{@{}p{\textwidth}@{}}{
\begin{minipage}[t]{0.16\textwidth}
\vspace{0pt}
\centering
\includegraphics[width=\linewidth]{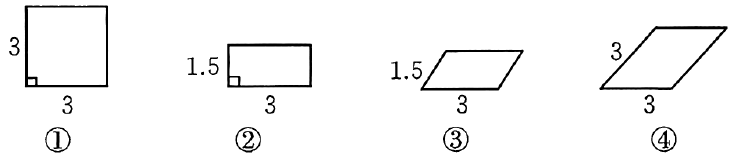}
\end{minipage}\hfill
\begin{minipage}[t]{0.81\textwidth}
\vspace{0pt}
\textbf{Question:} Using the oblique projection method to draw the perspective view of a square with a side length of $3\,\mathrm{cm}$, the correct one is \_ (fill in the number).\\
\textbf{Ground Truth:} $3$.
\end{minipage}
} \\
\midrule

Base Model
&
\begin{minipage}[t]{\linewidth}
\vspace{0pt}
\ttfamily\raggedright
<think>

In oblique projection, the front edge should keep the true length, so we prefer the option that preserves $3\,\mathrm{cm}$ on the visible edge.

1) \textbf{Option 1} is a square, so it does not look like a projected view.

2) \textbf{Option 2} uses $1.5\,\mathrm{cm}$, which seems too short for a $3\,\mathrm{cm}$ square.

3) \textbf{Option 3} also shows $1.5\,\mathrm{cm}$, so it is unlikely.

4) \textbf{Option 4} keeps the side length as $3\,\mathrm{cm}$ and looks like a slanted view, so choose it.

</think>

So the answer is option $\boxed{4}$.
\end{minipage}
& $\boxed{4}$ & No \\
\midrule

GRPO
&
\begin{minipage}[t]{\linewidth}
\vspace{0pt}
\ttfamily\raggedright
<think>

Oblique projection often shortens one direction. The rectangle with $1.5\,\mathrm{cm}\times 3\,\mathrm{cm}$ looks plausible as a shortened view, so choose option $\boxed{2}$.

</think>
\end{minipage}
& $\boxed{2}$ & No \\
\midrule

MBPO
&
\begin{minipage}[t]{\linewidth}
\vspace{0pt}
\ttfamily\raggedright
<think>

To solve this problem, I should verify both the projected shape and the length rule in oblique projection.

<look>

First identify which option actually shows a slanted receding direction, rather than a front-facing square or a plain rectangle.

</look>

1) \textbf{Option 1} is a front-facing square, so it is not an oblique projection.

2) \textbf{Option 2} is a rectangle and does not show the expected slanted receding direction.

<look>

Now check the side-length rule. In oblique projection of a square with side length $3\,\mathrm{cm}$, the front edge keeps its true length, while the receding edge is drawn shorter.

</look>

3) \textbf{Option 3} is slanted, keeps the front edge as $3\,\mathrm{cm}$, and uses $1.5\,\mathrm{cm}$ for the receding edge, which matches the expected oblique projection.

4) \textbf{Option 4} keeps both visible edges as $3\,\mathrm{cm}$, so it does not match the expected drawing rule.

</think>

So the answer is option $\boxed{3}$.
\end{minipage}
& $\boxed{3}$ & Yes \\
\bottomrule
\end{tabularx}
\end{table*}

\clearpage

\bibliographystyle{ACM-Reference-Format}
\bibliography{ACMMM/mbpo}